\documentclass{tlp}

\usepackage{amssymb}
\usepackage{psfig}
\usepackage[dvips]{graphicx}
\usepackage{aopmath}
\usepackage{times}
\usepackage{url}
\usepackage{latexsym}
\usepackage{exscale}
\usepackage{xspace} 
\usepackage{epsfig}
\usepackage[english]{varioref}
\usepackage[english]{babel}
\usepackage{theorem}
\usepackage{proof}

\newcommand {\qparl} {\mbox{$ [ $}}
\newcommand {\qparr} {\mbox{$ ] $}}

\newcommand {\subst} [3] {{#1}\qparl{#2}/{#3}\qparr}

\newcount\PLv\newcount\PLw\newcount\PLx\newcount\PLy\newdimen\PLyy\newdimen\PLX
\newbox\PLdot \setbox\PLdot\hbox{\tiny.} \def\scl{.08} 
\def\PLot#1{\PLx`#1\advance\PLx-42\PLy\PLx\PLv\PLx\divide\PLy9\PLw\PLy\multiply
\PLw9\advance\PLx-\PLw\advance\PLx-4\PLy-\PLy\advance\PLy4\PLX=\the\PLx pt
\advance\PLyy\the\PLy pt\wd\PLdot=\scl\PLX\raise\scl\PLyy\copy\PLdot}
\def\draw#1{\ifx#1\end\let\next=\relax\else\PLot#1\let\next=\draw\fi\next}

\def\invamp{\hbox{\PLyy=70pt\draw :::;DMV_gqppyyyyyooooxxxnnwvlutkjaWNE=5-./9%
9:::CCCC:::99/..--544=EENWWaajjjkktttttttNNNVVVVVVVV\end \hskip4pt}}
\newbox\iabox\setbox\iabox\invamp \def\Invamp{\copy\iabox}

\newcommand {\lneg} [1] {{#1}^\bot}

\newcommand {\dedLL} {\vdash}

\newcommand {\dedll} [2] {\dedLL {#1}:{#2}}
\newcommand {\dedllT} [1] {\dedll {\Theta}{#1}}

\newcount\PLv\newcount\PLw\newcount\PLx\newcount\PLy\newdimen\PLyy\newdimen\PLX
\newbox\PLdot \setbox\PLdot\hbox{\tiny.} 
\def\scl{.08}
\def\bscl{.095}
\def\lscl{.05}
\def\PLot#1{\PLx`#1\advance\PLx-42\PLy\PLx\PLv\PLx\divide\PLy9\PLw\PLy\multiply
\PLw9\advance\PLx-\PLw\advance\PLx-4\PLy-\PLy\advance\PLy4\PLX=\the\PLx pt
\advance\PLyy\the\PLy pt\wd\PLdot=\scl\PLX\raise\scl\PLyy\copy\PLdot}
\def\LPLot#1{\PLx`#1\advance\PLx-42\PLy\PLx\PLv\PLx\divide\PLy9\PLw\PLy\multiply\PLw9\advance\PLx-\PLw\advance\PLx-4\PLy-\PLy\advance\PLy4\PLX=\the\PLx pt
\advance\PLyy\the\PLy pt\wd\PLdot=\lscl\PLX\raise\lscl\PLyy\copy\PLdot}
\def\BPLot#1{\PLx`#1\advance\PLx-42\PLy\PLx\PLv\PLx\divide\PLy9\PLw\PLy
\multiply\PLw9\advance\PLx-\PLw\advance\PLx-4\PLy-\PLy\advance\PLy4\PLX=\the\PLx pt\advance\PLyy\the\PLy pt\wd\PLdot=\bscl\PLX\raise\bscl\PLyy\copy\PLdot}
\def\draw#1{\ifx#1\end\let\next=\relax\else\PLot#1\let\next=\draw\fi\next}
\def\bdraw#1{\ifx#1\end\let\next=\relax\else\BPLot#1\let\next=\bdraw\fi\next}
\def\ldraw#1{\ifx#1\end\let\next=\relax\else\LPLot#1\let\next=\ldraw\fi\next}
\def\invamp{\hbox{\PLyy=70pt\draw :::;DMV_gqppyyyyyooooxxxnnwvlutkjaWNE=5-./9%
9:::CCCC:::99/..--544=EENWWaajjjkktttttttNNNVVVVVVVV\end \hskip4pt}}
\def\Binvamp{\hbox{\PLyy=70pt\bdraw :::;DMV_gqppyyyyyooooxxxnnwvlutkjaWNE=5-./9%
9:::CCCC:::99/..--544=EENWWaajjjkktttttttNNNVVVVVVVV\end \hskip4pt}}
\def\Linvamp{\hbox{\PLyy=70pt\ldraw :::;DMV_gqppyyyyyooooxxxnnwvlutkjaWNE=5-./9%
9:::CCCC:::99/..--544=EENWWaajjjkktttttttNNNVVVVVVVV\end \hskip4pt}}
\newbox\iabox\setbox\iabox\invamp \def\Invamp{\copy\iabox}
\newbox\Biabox\setbox\Biabox\Binvamp \def\BInvamp{\copy\Biabox}
\newbox\Liabox\setbox\Liabox\Linvamp \def\LInvamp{\copy\Liabox}

\newcommand{\comment} [1]{}

\newcommand{\lprolog}{\textsf{$\lambda$Prolog}}

\newcommand{\caselp}{\textsf{CaseLP}}

\newcommand{\impact}{\textsf{IMPACT}}

\newcommand{\java}{\textsf{Java}}

\newcommand{\kqml}{\textsf{KQML}}

\newcommand{\prolog}{\textsf{Prolog}}
\newcommand{\sicstus}{\textsf{SICStus Prolog}}

\newcommand{\ehhf}{${\cal E}_{hhf}$}
\newcommand{\forum}{\textsf{Forum}}

\newcommand{\arpeggio}{ARPEGGIO}

\newcommand{\dylog}{\textsf{Dylog}}

\newcommand{\concMTM}{Concurrent \textsf{METATEM}}
\newcommand{\MTM}{\textsf{METATEM}}
\newcommand{\Lsitc}{${\cal L}_{sit-calc}$}
\newcommand{\congo}{\textsf{ConGolog}}
\newcommand{\golog}{\textsf{Golog}}

\newcommand{\agento}{\textsf{AGENT-0}}

\newcommand{\A}      {A}

\newcommand{\D}      {{\cal D}}

\newcommand{\G}      {{\cal G}}
\newcommand{\Head}   {{\cal H}}

\newcommand{\R}      {{\cal R}}

\newcommand{\scode}{\ensuremath{\protect\mathcal{C}}\xspace}
\newcommand{\sctype}{\ensuremath{\protect\mathcal{T}}\xspace}     
\newcommand{\scfunct}{\ensuremath{\protect\mathcal{F}}\xspace}    

\newcommand{\vect}{\ensuremath{{\vec t}}}

\newcommand{\pre}[1]{\ensuremath{\mathit{Pre}{\tt(#1)}}\xspace} 
\newcommand{\add}[1]{\ensuremath{\mathit{Add}{\tt(#1)}}\xspace} 
\newcommand{\del}[1]{\ensuremath{\mathit{Del}{\tt(#1)}}\xspace} 

\newcommand{\concur}{\ensuremath{\textbf{conc}}\xspace}

\newcommand{\bfF}{\ensuremath{\protect\mathbf{F}}\xspace}
\newcommand{\bfP}{\ensuremath{\protect\mathbf{P}}\xspace}
\newcommand{\bfW}{\ensuremath{\protect\mathbf{W}}\space}
\newcommand{\bfO}{\ensuremath{\protect\mathbf{O}}\xspace}
\newcommand{\bfDo}{\ensuremath{\protect\mathbf{Do\,}}\xspace}

\newcommand{\intcons}{\ensuremath{\protect\mathcal{IC}}\xspace}   

\newcommand{\agprog}{\ensuremath{\protect\mathcal{P}}\xspace}

\newcommand{\agstate}{\ensuremath{\protect\mathcal{O}}\xspace}
                                
\newcommand{\App}{\ensuremath{\textbf{App}_{\agprog,\agstate}(S)}\xspace}
\newcommand{\acl}[2]{\ensuremath{{\protect\mathbf{A}\textbf{-}\protect\mathbf{Cl}}_{#2}({#1})}\xspace}

\newcommand{\la}{\ensuremath{\leftarrow}\xspace}

\newcommand{\SC}{{\cal C}}
\newcommand{\Ob}{{\cal O}}
\newcommand{\OS}{\agstate}

\newcounter{myenumctr}

\newcommand{\IC}{{\mathcal IC}}

\newcommand{\cca}{{\tt cca}}
\newcommand{\ccall}{{\tt cc}}

 \newtheorem{definition}{Definition}

\newcommand{\intui}{\Rightarrow}
\newcommand{\into}{\Leftarrow}
\newcommand{\with}{\&}
\newcommand{\lolli}{\includegraphics[width=.3cm]{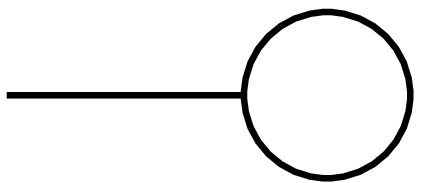}}
\newcommand{\lollo}{\includegraphics[width=.3cm]{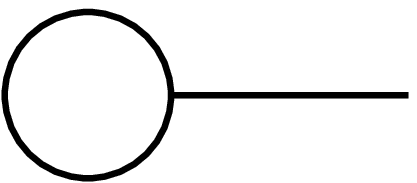}}

\newcommand{\para}{\mathrel{\Invamp}}
\newcommand{\bigpar}{\mathrel{\BInvamp}}
\newcommand{\littlepar}{\mathrel{\LInvamp}}
\newcommand{\tensor}{\otimes}
\newcommand{\anti}{\bot}
\newcommand{\all}{\top}
\newcommand{\one}{ {\bf 1} }

\newcommand{\poss}{$\lozenge$}
\newcommand{\nec}{$\Box$}
\newcommand{\nextt}{$\bigcirc$}
\newcommand{\nexttm}{\bigcirc}
\newcommand{\until}{$\cal U$}
\newcommand{\since}{$\cal S$}

\newcommand{\necm}{\square}
\newcommand{\stronglast}{\includegraphics[width=.3cm]{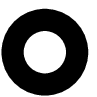}}
\newcommand{\weaklast}{\includegraphics[width=.3cm]{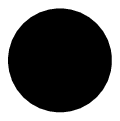}}
\newcommand{\was}{\blacklozenge} 
\newcommand{\heretofore}{\blacksquare}

\newcommand{\statecons}      {\Sigma_{\R}}
\newcommand{\stateatoms}     {\Pi_{\statecons}}

\newcommand{\munion}{\uplus}
\newcommand{\bigmunion}{\biguplus}

\newcommand{\forumseq}[4]{#1:#2; #3 \rightarrow #4}
\newcommand{\sforumseq}[3]{\forumseq{\Sigma}{#1}{#2}{#3}}

\newcommand{\rall}{halt}
\newcommand{\ranti}{erase}
\newcommand{\rpar}{sync}
\newcommand{\rwith}{and}
\newcommand{\rimpl}{fire}
\newcommand{\rintu}{augment}
\newcommand{\rforall}{hide}
\newcommand{\rress}{bc}
\newcommand{\rver}{verify}
\newcommand{\rmove}{move}

\newcommand{\rempty}{empty}

\newcommand{\instances}[1]{\langle #1\rangle} 

\newcommand{\arrowupdown}[4]{\setbox0=\hbox{$\ {}^{#2}\ $}
  \setbox1=\hbox{$\longrightarrow$}
  \ifdim\wd0<\wd1\setbox0=\box1\else\relax\fi
  {#1}\,\mathop{\hbox to \wd0{\rightarrowfill}}\limits^{#2}_{#3}\,{#4}
}

\newcommand{\arrowdown}[3]{\setbox0=\hbox{$\ {}_{#2}\ $}
  \setbox1=\hbox{$\longrightarrow$}
  \ifdim\wd0<\wd1\setbox0=\box1\else\relax\fi
  {#1}\,\mathop{\hbox to \wd0{\rightarrowfill}}\limits_{#2}\,{#3}
}

\newcommand\Arrow[3]{\setbox0=\hbox{$\ {}^{#2}\ $}
  \setbox1=\hbox{$\Rightarrow$}
  \loop\setbox1=\hbox{=\kern-0.3em\unhbox1}\ifdim\wd1<\wd0\repeat
  \hbox{${#1}\,\mathop{\box1}\limits^{#2\,}{#3}$}
}

\newcommand{\rhssep}{\mid\!\mid}
\newcommand{\sfoseq}[5]{#2;#3\rightarrow_{#1} #4\,\rhssep\,#5}
\newcommand{\foseq}[4]{\sfoseq{\Sigma}{#1}{#2}{#3}{#4}}

\newcommand{\foseqnt}[5]{{#1};{#2}[{#3}]\rightarrow_{\Sigma}{#4\,\rhssep\,#5}}
\newcommand{\sfoseqnt}[6]{{#2};{#3}[{#4}]\rightarrow_{#1}{#5\,\rhssep\,#6}}

\newcommand{\conc}{\parallel}

\newcommand{\calA}      {{\cal A}}
\newcommand{\calP}      {{\cal P}}
\newcommand{\calS}      {{\cal S}}
\newcommand{\calL}      {{\cal L}}

\begin{document}
\bibliographystyle{acmtrans}

\title{Logic-Based Specification Languages\\ 
for Intelligent Software Agents\thanks{Partially supported by the 
``Verifica di Sistemi Reattivi Basati su Vincoli (COVER)'' project of 
the Programma di Ricerca Cofinanziato MIUR, Bando 2002, and by the 
``Discovery'' project of the Australian Research Council number DP0209027.}}
\shorttitle{Logic-Based Specification Languages 
for Intelligent Software Agents}
\author[V. Mascardi, M. Martelli and L. Sterling]
{VIVIANA MASCARDI\\
DISI, Universit\`a di Genova,\\
Via Dodecaneso 35, 16146, Genova, Italy. \\
\email{mascardi@disi.unige.it}
\and
MAURIZIO MARTELLI\\
DISI, Universit\`a di Genova,\\
Via Dodecaneso 35, 16146, Genova, Italy. \\
\email{martelli@disi.unige.it}
\and
LEON STERLING\\
Department of Computer Science and Software Engineering, \\
The University of Melbourne \\
Victoria 3010, Australia.\\
\email{leon@cs.mu.oz.au}
}
\pagerange{\pageref{firstpage}--\pageref{lastpage}}
\volume{\textbf{XX} (XX):}
\jdate{XXX}
\setcounter{page}{1}
\pubyear{XXXX}

\maketitle
\label{firstpage}

\begin{abstract}
The research field of 
Agent-Oriented Software Engineering (AOSE) aims to find 
abstractions, languages, methodologies and toolkits for modeling, 
verifying, validating and prototyping complex  applications conceptualized as
Multiagent Systems (MASs).  A very lively research sub-field 
studies how formal methods can be used for AOSE. 
This paper presents a detailed survey of
six logic-based executable agent specification
languages that have been chosen for their potential to be integrated in our 
\arpeggio\ project, an open framework for specifying 
and prototyping a MAS. The six languages are 
\congo, \agento, the \impact\ agent programming language, \dylog,
\concMTM\ and \ehhf. For each executable language, the logic foundations
are described and an example of use is shown. A comparison
of the six languages and a survey of similar approaches complete the paper, together
with considerations of the advantages of using logic-based languages in MAS modeling and
prototyping. 
\end{abstract}

\begin{keywords}
 agent-oriented software engineering, logic-based language, multiagent system
\end{keywords}

\section{Introduction}

Today's software applications are typically extremely complex.
They may involve heterogeneous components which need to 
represent their knowledge 
about the world, about themselves, and about the other entities that 
populate the world, in order to reason about the world, to plan
future actions which
should be taken to reach some final goal and to take rapid 
decisions when the situation demands a quick reaction. Since knowledge
and competencies are usually distributed, the components need
to interact to exchange information or to delegate tasks. This interaction 
may follow sophisticated communication protocols. 
Due to component and system complexity, applications of this kind are difficult to be 
correctly and efficiently engineered. Indeed  a 
very active research area has been working for almost twenty years 
finding abstractions, languages, methodologies and toolkits for modeling, 
verifying, validating and finally implementing applications of this kind.

The underlying metaphor is that the components of complex real-world
applications are {\em intelligent agents}. The agents 
interact, exchanging information and collaborating for reaching
a common target, or compete to control some shared resource and 
to maximize their personal profit, building, in both cases, a society of agents, 
or {\em multiagent system} (MAS). 

An {\em intelligent agent}, according to a classical definition proposed 
by Jennings, Sycara and Wooldridge in \cite{jennings:98}, is
\begin{quote}
``a computer system, {\em situated} in some environment, that is
capable of {\em flexible autonomous} actions in order to meet its design
objectives.\\
Situatedness means that the agent receives sensory input
from its environment and that it can perform actions which change the
environment in
some way.

By  autonomy we mean that the system should be able to act
without the direct intervention of humans (or other agents), 
and that it should have control over its own actions and internal
state. [\ldots]
By flexible, we mean that the system is:
\begin{itemize}
\item  responsive: agents should perceive their environment and respond in a
timely
fashion to changes that occur in it;
\item  pro-active: agents should be able to exhibit opportunistic,
goal-directed
behavior and take the initiative when appropriate;
\item  social: agents should be able to interact, when appropriate, with other
artificial agents and humans.''
\end{itemize}
\end{quote}
Research on {\em agent-oriented software engineering} (AOSE)
\cite{petrie00agentbased,AOSE-Wooldridge-Ciancarini} 
aims at providing the means for engineering applications conceptualized as MASs.
As pointed out in \cite{AOSE-Wooldridge-Ciancarini}, 
the use of {\em formal methods} is one of the most active areas in this field, 
where formal methods play three roles:
\begin{itemize}
\item in the {\em specification} of systems;
\item for {\em directly programming} systems; and
\item in the {\em verification} of systems.
\end{itemize}
We think that logic-based formal methods can be very effective for 
fitting all three roles. In fact,  
the current predominant approach to specifying agents has involved treating the
agents as 
{\em intentional systems} that may be  understood by attributing 
to them {\em mental states}
 such as beliefs, desires and intentions \cite{dennett87intentional,WJ95,Woold2000}. 
A number of approaches for formally specifying agents 
as intention systems have been 
developed, capable of representing {\em beliefs, goals} and
{\em actions} of agents and the {\em ongoing interaction} among them.
A large number of logics appear  successful at formalizing these concepts
in a very intuitive and natural way, including for example 
 {\em modal logic}, {\em temporal 
logic} and {\em deontic logic}. 

Further, there are various logic-based languages for which a working
interpreter or an automatic mechanism for animating specifications exists. 
When these languages are used to specify agents, a working prototype of
the given specification is obtained for free and can be used for early testing and
debugging of specification. Most of the
executable logic-based languages suffer from significant limitations (very low efficiency,
poor scalability and modularity, no support for physical distribution of
the computation nor for integration of external packages and languages) which make them
only suitable for building simple prototypes. Nevertheless, even if these
languages will never be used to build the final application, their execution can give 
useful and very quick feedback to the MAS developer, who can take advantage of this early
testing and debugging for iteratively refining the MAS specification. 
 
Additionally, verification is the process of showing that an 
implemented system is correct with respect to its original specification. 
If the language in which the system is implemented is {\em axiomatizable}, 
deductive (axiomatic) verification is possible.
Otherwise, the {\em model checking} semantic approach can be followed: given a formula 
$\varphi$ of a logic $L$ and a model ${\cal M}$ for $L$, determine whether or not
 ${\cal M} \models_L \varphi$. There are logic-based
languages which have been axiomatized, allowing an axiomatic
verification, and other languages which can be used for model checking.

Logic-based formalisms are suitable for the stages of specification, 
direct execution and verification of MAS prototypes. We could ask if some 
environment and methodology exist that provide a set of logic-based languages for iteratively
facing these stages inside a common framework, until a working prototype 
that behaves as expected is obtained. A preliminary answer can come from 
\arpeggio. \arpeggio\ (\emph{Agent based Rapid Prototyping Environment Good for Global
Information Organization} \cite{arpeggio,zini00,mascardi02}) is an open framework 
where specification, 
execution and verification of MAS prototypes can be carried on choosing the most 
suitable language or languages from a set of supported ones.
\comment{
\footnote{At the 
beginning of the project, the languages that the environment aimed at including
were only logic-based ones. In more recent times we relaxed this 
constraint and also considered non-logic languages. In this paper
 we only refer to logic-based formalisms.}.
}

The rationale behind \arpeggio\ is that MAS development requires engineering support 
for a diverse range of software quality attributes. 
It is not feasible to create one monolithic AOSE approach 
to support all quality attributes.
Instead, we expect that different approaches will prove suitable to model,
verify, or implement different quality attributes. By providing the MAS developer
with a large set of languages and allowing the selection of the right language to
model, verify or implement each quality attribute, \arpeggio\ goes 
towards a modular approach to AOSE \cite{AAMAS03,creatingReusing}. 
\arpeggio\ is conceived as the framework
providing the building blocks for the development of an hybrid,
customizable AOSE methodology. It is not conceived as an hybrid system.

\arpeggio\ draws from three international logic programming
research groups: the Logic Programming Group at the Computer 
Science Department of the University of
Maryland, USA; the Logic Programming and Software Engineering 
Group at the Computer Science \&
Software Engineering Department of the University  of Melbourne, Australia; 
and the Logic Programming
Group at the Computer Science Department of the University of Genova, Italy.
An instance of the \arpeggio\ framework, \caselp\ 
\cite{MMZ99a,MMZ99b,MMMZ00,zini00,mascardi02}, has been developed
and tested on different real-world applications.
\comment{
 ranging 
from freight train traffic, in collaboration with FS s.p.a., the Italian railway 
company \cite{CGMMZ99a,CGMMZ99b}, to distributed health 
care \cite{Persano98}, from distributed database
transactions \cite{MM98}, to vehicle monitoring \cite{AMM00}, just to cite some of them. 
}
\caselp\ provides a language based on linear-logic, \ehhf\ \cite{Delzanno97,DM99},  
to specify and verify  agent specifications. This language, described 
in Section \ref{ehhf},
can also be executed, allowing the direct programming of the code for the
prototypical agents. Besides this high-level language, \caselp\ provides
an extension of \prolog\ for directly developing the MAS prototype. Although 
\caselp\ demonstrates that the concepts underlying the \arpeggio\
framework can be put into practice and can give interesting results,
the set of languages it provides is quite limited. Our motivation behind
the development of \arpeggio\ is to provide a broader set of languages so that the prototype
developer can choose the most suitable ones to model and/or program different
features of a MAS.

There are two main purposes of this paper. The first is to analyze a set of  
logic-based languages which have proven useful to specify, execute and/or 
validate agents and MAS prototypes and which have been integrated or 
could be integrated into the \arpeggio\ framework.
The set we have chosen consists of  \congo\ \cite{DLL}, \agento\ \cite{shoham93}, 
the \impact\ agent language \cite{eite-etal-98o}, \dylog\ \cite{BGMP}, 
\concMTM\ \cite{concMTM14} and \ehhf\ \cite{DM99}. These languages have been chosen because,
consistent with the \arpeggio\ philosophy,  a working
interpreter exists for them and they provide useful features
for specifying agent systems. Other languages possess these
features (see Section \ref{related-and-future}) and could have been chosen 
for being analyzed in this paper and for a future integration in \arpeggio.
However we preferred to provide a focused 
survey of a small subset of languages rather than a superficial 
description of a large set. 
In order to reach a real understanding of the main features
of the languages described in this paper
we have developed a common running example in all of them. 

The second purpose of this paper is to describe the different logics and calculi
the executable languages we take into consideration are based on, in order to 
provide a comprehensive survey on formalisms suitable to model intelligent 
agents.
Some executable languages are based on more than one logic; for example \concMTM\ 
is based on modal and temporal logic, and \textsf{AGENT-0} is based on modal and 
deontic logic. The classification we give of agent languages takes into 
account the predominant logic upon which the language is based.

The structure of the paper is the following:
\begin{itemize}
\item Section \ref{running-example} describes the running example we will
use throughout this paper to practically exemplify the features of the 
languages we will analyze;
\item Section  \ref{sit-calculus} introduces the situation calculus and 
the \congo\ agent programming language based on it;
\item Section  \ref{modal-logic} discusses modal logic and the 
\agento\ language;
\item in Section \ref{deontic-logic} the main features of deontic logic are shown and
the \impact\ programming language is analyzed as an example of an agent language
including deontic operators;
\item  Section  \ref{dynamic-logic} discusses dynamic logic and the 
\dylog\ agent language based on it;
\item Section \ref{temporal-logic} describes temporal logic and the \concMTM\
agent programming language;
\item Section \ref{linear-logic} introduces linear logic and 
analyses the \ehhf\ language included in the \caselp\ instance of the
\arpeggio\ framework;
\item Section \ref{spec-logic-comp} compares the agent programming languages 
introduced so far based on a set of relevant AOSE features they can support;
\item Section \ref{related-and-future} discusses related work;
\item finally, Section \ref{spec-logic-conc} concludes the paper. 
\end{itemize}

\section{The Running Example}
\label{running-example}

\begin{figure}[]
\begin{center}
\includegraphics[height=6.5cm]{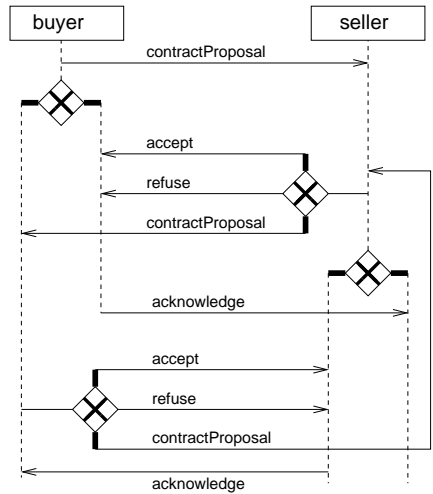}
\end{center}
\caption{The contract proposal protocol.}
\label{contractProtocol}
\end{figure}

To show how to define an agent in the various agent languages we discuss 
in this paper, we use a simple example of a seller agent in a distributed 
marketplace  which follows the communication 
protocol depicted in Figure \ref{contractProtocol}. The notation used in this 
figure is based on an agent-oriented extension of \textsf{UML} \cite{Parunak-odell,aose54};
the diamond with the  $\times$ inside represents a ``xor'' connector and the protocol 
can be repeated more than once (note the bottom arrow from the buyer to the 
seller labeled with a ``contractProposal'' message, which loops around and
up to point higher up to the seller time line). 

The seller agent
may receive a \textsf{contractProposal} message from a buyer agent.
According to the amount of merchandise required and the price proposed by the buyer, 
the seller may accept the proposal, refuse it or try to negotiate a 
new price by sending a \textsf{contractProposal} message back to the buyer.
The buyer agent can do the same (accept, refuse or negotiate) 
when it receives a \textsf{contractProposal} message back from the seller.

The rules guiding the behavior of the seller agent are the following:
\begin{itemize}
\item[] if the received message is 
\textsf{contractProposal(merchandise, amount, proposed-price)} then
\begin{itemize}
\item if there is enough merchandise in the warehouse and the price is greater or
equal than a 
\textsf{max} value, the seller accepts the proposal by sending an 
\textsf{accept} message to the buyer and 
concurrently ships the 
required merchandise to the buyer (if it is not possible to define 
concurrent actions, answering and  shipping merchandise will be executed 
sequentially);
\item if there is not enough merchandise in the warehouse or the price is lower 
or equal than
a \textsf{min} value, the seller agent refuses the proposal by sending a \textsf{refuse}
 message to the buyer;
\item if there is enough merchandise in the warehouse and the price is 
between \textsf{min} and \textsf{max}, the seller sends a 
\textsf{contractProposal} to the buyer
with a proposed price evaluated as the mean of the price proposed by 
the buyer and \textsf{max} (we will sometimes omit the definition of this
function, which is not of central interest in our example) .
\end{itemize} 
\end{itemize}
In our example, the merchandise to be exchanged are oranges, with minimum and maximum
price 1 and 2 euro respectively. The initial amount of oranges that the seller
possesses is 1000. 

Our example involves features which fall in the intersection of the
six languages and it is therefore quite simple. 
An alternative choice to providing a simple unifying example
would consist of providing six sophisticated examples highlighting
the distinguishing features of each of the six languages.
However, while sophisticated ad-hoc examples can be found in the papers discussing
the six languages, a unifying (though simple) example had not been 
proposed yet. 
Consistent with the introductory nature of our paper 
and with the desire to contribute in an original way to the
understanding of the six languages, we 
opted for the simple unifying example, which is both introductory 
and original. 
The about seventy references included in the following six sections 
should help the reader in finding all the documents 
she/he needs for deepening her/his knowledge about the six languages
discussed in this paper. 

\section{Situation Calculus}
\label{sit-calculus}

The situation calculus \cite{mcCarthy17} is well-known in AI research.
More recently there have been attempts 
to axiomatize it. The following description is based upon 
 \cite{PirriReiter24}. \Lsitc\ is a second order language with 
equality. It has three disjoint sorts: {\em action} for actions, {\em situation} for 
situations and a catch-all sort {\em object} for everything else depending on the domain 
of application. Apart from the standard alphabet of logical symbols ($\wedge, \neg$ and
 $\exists$, used with their usual meaning), \Lsitc\ has the following alphabet:
\begin{itemize}
\item Countably infinitely many individual variable symbols of each sort and
countably infinitely many predicate variables of all arities. 
\item Two function symbols of sort {\em situation}:
\begin{enumerate}
\item A constant symbol $S_0$, denoting the initial situation.
\item A binary function symbol 
$do:action  \times  situation  \rightarrow  situation$. \\
$do(a,s)$ denotes the successor 
situation resulting from performing action $a$  in situation $s$. 
\end{enumerate}
\item A binary predicate symbol $\sqsubset:situation  \times  situation$, defining an 
ordering relation on situations. The intended interpretation of situations is as 
action histories, in which case $s \sqsubset s'$ means that  $s'$ 
can be reached by $s$ by a finite application of actions.
\item A binary predicate symbol $Poss:action  \times  situation$. The intended 
interpretation of $Poss(a, s)$ is that it is possible to perform the action $a$ 
in the situation $s$.
\item Countably infinitely many predicate symbols used to denote 
situation independent relations and countably infinitely many function symbols 
used to denote situation independent functions.
\item A finite or countably infinite number of
 function symbols 
called {\em action functions} and used to denote actions.
\item A finite or countably infinite number of
{\em relational fluents} (predicate symbols used to denote situation dependent relations).
\item A finite or countably infinite number of
 function symbols called {\em functional fluents} and used to denote 
situation dependent functions.
\end{itemize}

In the axiomatization proposed in \cite{LPR98}, axioms are divided into
domain axioms and  domain independent
foundational axioms for situations. Besides axioms, 
\cite{LPR98} also introduces basic theories of actions and 
a metatheory for the situation calculus which allows to determine when a basic action
theory is satisfiable and when it entails a particular kind of sentence, 
called {\em regressable sentences}.   
Here we only discuss domain independent
foundational axioms for situations.
Since the scope of this paper is to provide introductory material which can be 
understood with little effort, we will address neither  domain axioms nor the
metatheory for the situation calculus, both of which require 
a strong technical background.   

\subsection{Foundational Axioms for Situations}

There are four foundational axioms for the situation calculus, based on 
\cite{PirriReiter24} but simpler that the ones presented there. 
They capture the
intuition that situations are finite sequences of actions where the 
second order
induction principle holds, and that there is
a ``subsequence'' relation among them. In the following axioms, $P$ is a 
predicate symbol.

\begin{equation}
\label{axSitc1}
do(a_1, s_1) = do(a_2, s_2) \Rightarrow a_1 = a_2 \wedge s_1 = s_2
\end{equation}
\begin{equation}
\label{axSitc2}
\forall P. P(S_0) \wedge \forall a, s. [P(s) \Rightarrow P(do(a, s))] \Rightarrow 
\forall s. P(s)
\end{equation}
Axiom \ref{axSitc1} is a unique name axiom for situations: two situations are the 
same iff they are the same sequence of actions.
Axiom \ref{axSitc2} is second order induction on situations.
The third and fourth axioms are:
\begin{equation}
\label{axSitc3}
\neg~(s \sqsubset S_0)
\end{equation}
\begin{equation}
\label{axSitc4}
(s \sqsubset do(a, s')) \equiv (s \sqsubseteq s')
\end{equation}
Here $s \sqsubseteq s'$ is an abbreviation for $(s \sqsubset s') \vee  (s = s')$.
The relation $\sqsubset$ provides an ordering relation on situations. Intuitively,
$s \sqsubset s'$ means that the situation $s'$ can be obtained from the 
situation $s$ by adding one or more actions to the end of $s$.

The above four axioms are {\em domain independent}. They provide the basic 
properties of situations in any domain specific axiomatization of particular 
fluents and actions.

\subsection{\congo}
\label{congo}

\congo\ is a concurrent programming language based on the situation calculus which 
includes facilities for prioritizing the concurrent execution, interrupting the execution
 when certain conditions become true, and dealing with exogenous actions. As stated 
by De Giacomo, Lesp\'erance and Levesque in \cite{DLL}, the adoption of a 
language like \congo\ is
a promising alternative to traditional plan synthesis, since it allows high-level 
program execution. \congo\ is an extension of the programming language 
\golog\ \cite{congo15}: in Section \ref{golog} we present  \golog\ and
 in Section \ref{congoBis} we deal with its extension \congo.

\subsubsection{\golog}
\label{golog}

\golog\ is a logic-programming language whose primitive actions are drawn from a 
background domain theory. 

\golog\ programs are inductively defined as: 
\begin{itemize}
\item Given a situation calculus 
action $a$ with all situation arguments in its parameters replaced by the special 
constant $now$, $a$ is a \golog\ program (primitive action). 
\item  Given a situation calculus 
formula $\varphi$ with all situation arguments in its parameters replaced by the special 
constant $now$, $\varphi?$ is a \golog\ program (wait for a condition).
\item  Given $\delta$, $\delta_1$, $\delta_2$, $\delta_n$ \golog\ programs,
	\begin{itemize}
	\item $(\delta_1 ; \delta_2)$ is a \golog\ program (sequence);
	\item $(\delta_1 \mid \delta_2)$ is a \golog\ program
	(nondeterministic choice between actions);
	\item $\pi v. \delta$ is a \golog\ program (nondeterministic choice of arguments);
	\item $\delta^*$ is a \golog\ program (nondeterministic iteration);
	\item \{{\bf proc} $P_1(\stackrel{\rightarrow}{v_1}) \delta_1$ {\bf end};
\ldots 
{\bf proc} $P_n(\stackrel{\rightarrow}{v_n}) \delta_n$ {\bf end}; $\delta$ \}
is a \golog\ program (procedure: $P_i$ are procedure names and $v_i$ are their
parameters).
	\end{itemize}
\end{itemize}

\comment{
It includes the following constructs:

\begin{tabbing}
Spaziolo \= Formulalala \= Spaziolo \=  \kill 
\> $a$ \>  \> Primitive action.\\
\> $\varphi?$ \> \> Wait for a condition.\\
\> $(\delta_1 ; \delta_2)$ \> \> Sequence.\\
\> $(\delta_1 \mid \delta_2)$ \> \> Nondeterministic choice between actions.\\
\> $\pi v. \delta$ \> \> Nondeterministic choice of arguments.\\
\> $\delta^*$ \> \> Nondeterministic iteration.\\
\end{tabbing}  
\vspace*{-.9cm}
\begin{tabbing}
Spaziolo \= Formulalalaaaaaaaaaaaaaaaaaaaaaaaaaaaaaaaaaaaa \= Spaziolo \=  \kill 
\> \{{\bf proc} $P_1(\stackrel{\rightarrow}{v_1}) \delta_1$ {\bf end};
\ldots 
{\bf proc} $P_n(\stackrel{\rightarrow}{v_n}) \delta_n$ {\bf end}; $\delta$ \} \> \>
Procedures.
\end{tabbing}  
In the first line, $a$ stands for a situation calculus action 
with all situation arguments in its parameters replaced by the special 
constant {\em now}. Similarly in the line below $\varphi$ stands for 
a situation calculus formula with all situation arguments replaced by {\em now}. 
In the following lines, $\delta$, possibly 
subscripted, ranges over \golog\ programs. 
}

\paragraph*{Program Execution.} Given a domain theory ${\cal D}$ and a program $\delta$ 
the execution task is to find a sequence  $\stackrel{\rightarrow}{a}$ of actions such that:
\[ {\cal D} \models Do(\delta, S_0, do(\stackrel{\rightarrow}{a}, S_0)) \]
where 
\\
\centerline{$Do(\delta, s, s')$}
means that program $\delta$ when executed starting in situation $s$ has $s'$ as a 
legal terminating situation, and 
\\
\centerline{$do(\stackrel{\rightarrow}{a}, s) = do([a_1, \ldots, a_n], s)$}
 is an abbreviation for 
\\
\centerline{
$do(a_n, do(a_{n-1}, \ldots, do(a_1, s)))$.}
Since \golog\ programs can be nondeterministic, there may be several terminating situations for the same program and starting situation. 
$Do(\delta, s, s')$ is formally defined by means of the following inductive definition:
\begin{enumerate}
\item Primitive actions ($a[s]$ denotes the action 
obtained by substituting the situation variable $s$ for all occurrences of $now$ 
in functional fluents appearing in $a$):
\[ Do(a, s, s') \stackrel{def}{=} Poss(a[s], s) \wedge s' = do(a[s], s) \]

\item Wait/test actions ($\varphi[s]$ denotes the formula 
obtained by substituting the situation variable $s$ for all occurrences of $now$ 
in functional and predicate fluents appearing in $\varphi$):
\[ Do(\varphi ?, s, s') \stackrel{def}{=} \varphi [s] \wedge s = s' \]
\item Sequence:
\[ Do(\delta_1;\delta_2, s, s') \stackrel{def}{=} \exists s''. 
Do(\delta_1, s, s'') \wedge Do(\delta_2, s'', s')\]
\item Nondeterministic branch:
\[ Do(\delta_1 \mid \delta_2, s, s') \stackrel{def}{=} 
Do(\delta_1, s, s') \vee Do(\delta_2, s, s')\]
\item Nondeterministic choice of argument ($\pi x.\delta(x)$ is executed
by nondeterministically picking an individual $x$, and for that $x$, performing the
program  $\delta(x)$):
\[ Do(\pi x.\delta(x), s, s') \stackrel{def}{=} \exists x. 
Do(\delta(x), s, s') \]
\item Nondeterministic iteration:
\\
\\
\centerline{ 
$ Do(\delta^*, s, s') \stackrel{def}{=} 
\forall P. \{ \forall s_1. P(s_1, s_1) \wedge \forall s_1, s_2, s_3. [P(s_1, s_2) \wedge Do(\delta, s_2, s_3)$} 
\\
\centerline{$
\Rightarrow P(s_1, s_3) ] \} \Rightarrow P(s, s')$}
\\
$P$ is a binary predicate symbol. Saying ``$(x, x')$ is in the set
(defined by $P$)'' is equivalent to saying ``$P(x, x')$ is true''.
Doing action $\delta$ zero or more times leads from the situation $s$ to the situation $s'$ 
if and only if $(s, s')$ is in every set (and therefore, the smallest set) such that:
\begin{enumerate}
\item $(s_1, s_1)$ is in the set for all situations $s_1$.
\item Whenever $(s_1, s_2)$ is in the set, and doing $\delta$ in situation $s_2$ 
leads to situation $s_3$, then $(s_1, s_3)$ is in the set. 
\end{enumerate}   
The above is the standard second order definition of the set 
obtained by nondeterministic iteration.  
\end{enumerate}
We do not deal with expansion of procedures. 
The reader can see 
\cite{congo15} for the details.

\subsubsection{\congo}
\label{congoBis}

\congo\ is an extended version of \golog\ that incorporates 
concurrency, handling:
\begin{itemize}
\item concurrent processes with possibly different priorities;
\item high-level interrupts and
\item arbitrary exogenous actions.
\end{itemize}

\congo\ programs are defined by the following inductive rules:
\comment{
\begin{tabbing}
Spaziolo \= Formulalalalalala \= Spaziolo \=  \kill 
\> {\bf if} $\varphi$ {\bf then} $\delta_1$ {\bf else} $\delta_2$ \>  
\> Synchronized conditional.\\
\> {\bf while} $\varphi?$ {\bf do} $\delta$ \> \> Synchronized loop.\\
\> $(\delta_1 \parallel \delta_2)$ \> \> Concurrent execution.\\
\> $(\delta_1 \rangle \rangle \delta_2)$ \> \> Concurrency with different priorities.\\
\> $\delta^{\parallel}$ \> \> Concurrent iteration.\\
\> $\langle \varphi \rightarrow \delta \rangle$ \> \> Interrupt.\\
\end{tabbing}
}
\begin{itemize}
\item All \golog\ programs are \congo\ programs.
\item  Given a situation calculus 
formula $\varphi$ with all situation arguments in its parameters replaced by the special 
constant $now$, and $\delta$, $\delta_1$, $\delta_2$ \congo\ programs,
	\begin{itemize}
	\item {\bf if} $\varphi$ {\bf then} $\delta_1$ {\bf else} $\delta_2$ 
	is a \congo\ program (synchronized conditional);
	\item {\bf while} $\varphi?$ {\bf do} $\delta$ is a \congo\ program
	(synchronized loop);
	\item $(\delta_1 \parallel \delta_2)$  is a \congo\ program 
	(concurrent execution);
	\item $(\delta_1 \rangle \rangle \delta_2)$ is a \congo\ 
	program (concurrency with different priorities);
	\item $\delta^{\parallel}$ is a \congo\ program (concurrent iteration);
	\item $<$$\varphi \rightarrow \delta$$>$ is a \congo\ program (interrupt).
	\end{itemize} 
\end{itemize}

The constructs {\bf if} $\varphi$ {\bf then} $\delta_1$ {\bf else} $\delta_2$ and 
{\bf while} $\varphi?$ {\bf do} $\delta$ are the synchronized versions of the usual 
if-then-else and while-loop. They are synchronized in the sense that the test of the 
condition $\varphi$ does not involve a transition per se: the evaluation of the condition 
and the first action of the branch chosen will be executed as an atomic action.
 The construct $(\delta_1 \parallel \delta_2)$ denotes the concurrent execution of the
 actions $\delta_1$ and $\delta_2$. $(\delta_1 \rangle \rangle \delta_2)$ denotes the 
concurrent execution of the actions $\delta_1$ and $\delta_2$ with $\delta_1$ having 
higher priority than $\delta_2$, restricting the possible interleavings of the two 
processes: $\delta_2$ executes only when $\delta_1$ is either done or blocked.
The construct $\delta^{\parallel}$ is like nondeterministic iteration, but where the 
instances of $\delta$ are executed concurrently rather than in sequence. 
Finally, $\langle \varphi \rightarrow \delta \rangle$  
is an interrupt. It has two parts: a trigger 
condition $\varphi$ and a body $\delta$. The idea is that the body $\delta$ will execute 
some number of times. If $\varphi$ never becomes true, $\delta$ will not execute at all. 
If the interrupt gets control from higher priority processes when $\varphi$ is true, 
then $\delta$ will execute. Once it has completed its execution, the interrupt is 
ready to be triggered again. This means that a high priority interrupt can take complete 
control of the execution. 

\subsubsection{Semantics}

The semantics of \golog\ and \congo\ is in the style of transition semantics. 
Two predicates are defined 
which say when a program $\delta$ can legally terminate in a certain situation
$s$ ($Final(\delta, s)$) and when a program $\delta$ in the situation $s$ can legally
execute one step, ending in situation $s'$ with program $\delta'$ remaining 
($Trans(\delta, s, \delta', s')$). $Final$ and $Trans$ are characterized by
a set of equivalence axioms, each depending on the structure of the first argument. 
To give the flavor of how these axioms look like,
we show the ones for  empty program $nil$, 
atomic action $a$, testing $\varphi?$, nondeterministic branch $(\delta_1 \mid 
\delta_2)$
and concurrent execution  $(\delta_1 \parallel \delta_2)$\footnote{In order to define 
axioms properly,
programs should be encoded as first order terms.  We avoid dealing with this 
encoding, and describe axioms as if  
programs were already first order terms.}. The reader can find the complete set of axioms 
for $Final$ and $Trans$ in \cite{DLL}.

\begin{tabbing}
Spaziolo \= Formulalalalalalalalalalala \= *iff* \=  \kill 
\> $Trans(nil, s, \delta', s')$ \> $\equiv$ \> false \\ 
\> $Trans(a, s, \delta', s')$ \> $\equiv$ \> $Poss(a[s],s) \wedge \delta'
 = nil \wedge
s'= do(a[s],s)$ \\ 
\> $Trans(\varphi?, s, \delta', s')$ \> $\equiv$ \> $\varphi[s] \wedge \delta'
 = nil \wedge
s'= s$ \\ 
\> $Trans(\delta_1 \mid \delta_2, s, \delta', s')$ \> $\equiv$ \> 
$Trans(\delta_1, s, \delta', s') \vee 
Trans(\delta_2, s, \delta', s')$  \\ 
\> $Trans(\delta_1 \parallel \delta_2, s, \delta', s')$ \> $\equiv$ \> \\
\>  
$~~~~~~\exists \gamma. \delta' = (\gamma \parallel \delta_2) \wedge Trans(\delta_1, s, 
\gamma, s')
\vee$ \> \> \\
\> 
$~~~~~~\exists \gamma. \delta' = (\delta_1 \parallel \gamma) \wedge Trans(\delta_2, s,
\gamma, s')$ \> \> \\
\end{tabbing} 
The meaning of these axioms is that: $(nil, s)$ does not evolve to any configuration; 
$(a, s)$ evolves to $(nil, do(a[s],s))$ provided that $a[s]$ is possible in $s$;
$(\varphi?, s)$ evolves to $(nil, s)$ provided that $\varphi[s]$ holds; 
$(\delta_1 \mid \delta_2, s)$ can evolve to $(\delta', s')$ provided that either
$(\delta_1, s)$  or $(\delta_2, s)$  can do so; and finally, 
$(\delta_1 \parallel \delta_2, s)$ can evolve if $(\delta_1, s)$  can evolve and 
$\delta_2$ remains unchanged or $(\delta_2, s)$ 
can evolve and $\delta_1$ remains unchanged.

\begin{tabbing}
Spaziolo \= Formulalalalalalala \= *iff* \=  \kill 
\> $Final(nil, s)$ \> $\equiv$ \> true \\ 
\> $Final(a, s)$ \> $\equiv$ \> false \\ 
\> $Final(\varphi?, s)$ \> $\equiv$ \> false \\ 
\> $Final(\delta_1 \mid \delta_2, s)$ \> $\equiv$ \> 
$Final(\delta_1, s) \vee Final(\delta_2, s)$ \\ 
\> $Final(\delta_1 \parallel \delta_2, s)$ \> $\equiv$ \> 
$Final(\delta_1, s) \wedge Final(\delta_2, s)$ \\ 
\end{tabbing} 
These axioms say that $(nil,s)$  is a final configuration
while neither $(a,s)$ nor $(\varphi?,s)$ are.
$(\delta_1 \mid \delta_2, s)$ is final if either $(\delta_1, s)$ is or
$(\delta_2, s)$ is, while $(\delta_1 \parallel \delta_2, s)$ is final 
if both $(\delta_1, s)$ and
$(\delta_2, s)$ are.

The possible configurations that can be reached by a program $\delta$ in situation $s$ are
those obtained by repeatedly following the transition relation denoted by $Trans$
starting from $(\delta, s)$. The reflexive transitive closure of $Trans$ is denoted by
$Trans$*. By means of $Final$ and $Trans$* it is possible to give a new 
definition of $Do$ as
\[ Do(\delta, s, s') \stackrel{def}{=} \exists \delta'. Trans^{*} (\delta, s, \delta', s')
\wedge Final(\delta', s')\] 

\subsubsection{Implementation}

A simple implementation of \congo\ has been developed in \prolog. The definition of the
interpreter is lifted directly from the definitions of $Final$, $Trans$ and $Do$ 
given above. The interpreter requires that the program's precondition axioms, 
successor state axioms and axioms about the initial situation be expressible as 
\prolog\ clauses. In particular, the usual {\em closed world assumption} is made on the 
initial situation. Section 8 of \cite{DLL} describes the \congo\ interpreter in detail
and proves its correctness under suitable assumptions. The interpreter is included into 
a more sophisticated toolkit which provides facilities for debugging \congo\ programs
and delivering process modeling applications 
by means of a graphical interface.  
Visit \cite{cogrobo} to download
the interpreter of both \congo\ and its extensions discussed in the next section. 

\subsubsection{Extensions}

Some variants of \congo\ have been developed in the last years:
\begin{itemize}
\item \textsf{Legolog} ({\em LEGO MINDSTORM in (Con)Golog} \cite{legolog}) 
uses a controller from the 
\golog\ family of planners to control a MINDSTORM robot. 
\textsf{Legolog} is capable of dealing with primitive actions, 
exogenous actions and sensing; the \golog\ controller is replaced 
with an alternate planner. 
Visit \cite{legologURL} for details.

\item \textsf{IndiGolog} ({\em Incremental Deterministic (Con)Golog} \cite{Giuseppe02})
 is a high-level programming language 
where programs are executed incrementally to allow for interleaved action, 
planning, sensing, and exogenous events. \textsf{IndiGolog} provides a practical 
framework for real robots that must react to the environment and continuously 
gather new information from it. To account for planning, \textsf{IndiGolog} 
provides a local lookahead mechanism with a new language construct 
called {\em the search operator}.
\item \textsf{CASL} ({\em Cognitive Agent Specification Language} 
\cite{shapiro-lesperance-levesque}) is a framework for 
specifying complex MASs which also provides a verification environment based on
the \textsf{PVS} verification system \cite{shapiro-lesperance-levesque10}.
\item A class of knowledge-based \golog\ programs is extended 
with sense actions in \cite{Reiter01}. 
\end{itemize}

Most of the publications on \golog, \congo\ and their extensions can
be found at \cite{cogrobo}.

\subsubsection{Example}

The program for the seller agent, written in \congo, could look as
follows. The {\em emphasized} text is used for constructs of the language;
 the normal text is used for comments. Lowercase symbols represent constants
of the language and uppercase symbols are variables. Predicate and function
symbols are lowercase (thus, the {\em Poss} predicate symbol introduced in
the beginning of Section \ref{sit-calculus} is written
{\em poss} in the example). These Prolog-like conventions 
will be respected in all the examples appearing in the paper, 
unless stated otherwise. 
\begin{itemize}
\item Primitive actions declaration:
	\begin{itemize}
	\item[] {\em ship(Buyer, Merchandise, Required-amount)}\\
        The seller agent delivers the {\em Required-amount} of {\em Merchandise}
        to the {\em Buyer}.
	\item[] {\em send(Sender, Receiver, Message)}\\
        {\em Sender} sends {\em Message} to {\em Receiver}. 
	\end{itemize}
\item Situation independent functions declaration:
       \begin{itemize}
	\item[] {\em min-price(Merchandise) = Min}\\
        The minimum price the seller is willing to take under consideration
        for {\em Merchandise} is {\em Min}.
	\item[] {\em max-price(Merchandise) = Max}\\
         The price for {\em Merchandise} that the seller accepts without negotiation is equal or
         greater than {\em Max}.
	\end{itemize}
\item Primitive fluents declaration:
 \begin{itemize}
	\item[] {\em receiving(Sender, Receiver, Message, S)}\\
        {\em Receiver} receives {\em Message} from {\em Sender} in situation {\em S}.
	\item[] {\em storing(Merchandise, Amount, S)}\\
         The seller stores {\em Amount} of {\em Merchandise} in situation {\em S}.
	\end{itemize}
\item Initial situation axioms:
\begin{itemize}
\item[] {\em min-price(orange) = 1}
\item[] {\em max-price(orange) = 2}
\item[] $\forall$ {\em S, R, M}. $\neg$ {\em receiving(S, R, M, $s_0$)}
\item[] {\em storing(orange, 1000, $s_0$)}
\end{itemize}
\item Precondition axioms:
\begin{itemize}
	\item[] {\em poss(ship(Buyer, Merchandise, Required-amount), S) $\equiv$ \\
                 \hspace*{.5cm }storing(Merchandise, Amount, S) 
$\wedge$ Amount $\geq$ Required-amount}\\
               It is possible to ship merchandise iff there is enough merchandise stored in the warehouse.
	\item[] {\em poss(send(Sender, Receiver, Message), S) $\equiv$ true}\\
             It is always possible to send messages.
	\end{itemize}
\item Successor state axioms:
\begin{itemize}
	\item[] {\em receiving(Sender, Receiver, Message, \\
        \hspace*{.5cm} do(send(Sender, Receiver, Message), S)) $\equiv$ true}\\
         {\em Receiver} receives {\em Message} from 
         {\em Sender} in
         {\em do(send(Sender, Receiver, Message), S)}
         reached by executing {\em send(Sender, Receiver, Message)} in {\em S}. 
         For sake of conciseness we opted for a very simple formalization of 
         agent communication. More sophisticated formalizations can be found in 
         \cite{marcu95} and \cite{levesqueLNAI1441}. 

	\item[]
                 {\em storing(Merchandise, Amount, do(A, S)) $\equiv$\\
                  \hspace*{.5cm} (A = ship(Buyer, Merchandise, Required-amount) $\wedge$\\
                  \hspace*{.5cm} storing(Merchandise, Required-amount + Amount, S))\\
		  \hspace*{.5cm}	$\vee$ 
                  (A $\not=$ ship(Buyer, Merchandise, Required-amount) $\wedge$\\
                  \hspace*{.5cm} storing(Merchandise, Amount, S))}\\
       The seller has a certain {\em Amount} of {\em Merchandise} if it had 
      {\em Required-amount + Amount} of 
       {\em Merchandise} in the previous situation and it 
       shipped {\em Required-amount} of {\em Merchandise}, or if it had 
      {\em Amount} of 
       {\em Merchandise} in the previous situation and it 
       did not ship any {\em Merchandise}. 
	\end{itemize}
\end{itemize}
We may think that a buyer agent executes a {\em buyer-life-cycle} 
procedure concurrently with the seller agent procedure {\em seller-life-cycle}.
{\em buyer-life-cycle} defines the actions the buyer agent takes according to its
internal state and the messages it receives. The {\em seller-life-cycle} is defined
in the following way.
\\
\\
{\bf proc} {\em seller-life-cycle}
\\
\\
   \hspace*{.4cm}{\bf while} {\em true} {\bf do}
   \vspace*{-.2cm}
		\begin{itemize} 
		\item[] {\bf if} {\em receiving(Buyer, seller, 
                 contractProposal(Merchandise, Required-amount, Price), now)}
		\item[] {\bf then} 
			\begin{itemize} 
			\item[] {\bf if} {\em storing(Merchandise, Amount, now)\\
                                       $\wedge$ Amount $\geq$ Required-amount \\
                                       $\wedge$ Price $\geq$ max-price(Merchandise)}
			\item[] {\bf then} {\em ship(Buyer, Merchandise, Required-amount) \\
                                          $\conc$
                                          send(seller, Buyer, 
                                            accept(Merchandise, Required-amount, Price))}
			\item[] {\bf else}
 				\begin{itemize} 
				\item[] {\bf if} {\em (storing(Merchandise, Amount, now) \\
                                       $\wedge$ Amount $<$ Required-amount) \\ 
                                      $\vee$  Price $\leq$ min-price(Merchandise)}
				\item[] {\bf then} {\em send(seller, Buyer, 
                                            refuse(Merchandise, Required-amount, Price))}
				\item[] {\bf else}
					\begin{itemize} 
					\item[] {\bf if} 
                                        {\em storing(Merchandise, Amount, now) \\
                                        $\wedge$ Amount $\geq$ Required-amount\\
                                        $\wedge$ 
                                        min-price(Merchandise) $<$ Price $<$ max-price(Merchandise)}
					\item[] {\bf then} {\em send(seller, Buyer, \\
                                            contractProposal(Merchandise,
                                            Required-amount, \\
                                            (Price+max-price(Merchandise))/2))}
					\item[] {\bf else} {\bf nil}
                                        \end{itemize}
				\end{itemize}
			\end{itemize}
                \item[] {\bf else} {\bf nil}
		\end{itemize}
	

\section{Modal Logic}
\label{modal-logic}

This introduction is based on \cite{fisher-owens}.
Modal logic is an extension of classical logic with 
(generally) a new connective
\nec\ and its derivable counterpart \poss, known as {\em necessity} and
{\em possibility} respectively. If a formula \nec $p$ is true, it 
means that 
$p$ is necessarily true, i.e. true in every possible scenario, and \poss $p$
means that $p$ is possibly true, i.e. true in at least one possible scenario. 
It is possible to define \poss\ in terms of \nec:
\[ \lozenge p \Leftrightarrow \neg \Box \neg p \] 
so that $p$ is possible exactly when its negation is not necessarily true.
In order to give meaning to \nec\ and \poss, models for modal logic are usually
based on {\em possible worlds}, which are essentially a collection of connected
models for classical logic. The possible worlds are linked by a relation which
determines which worlds are accessible from any given world. 
It is this {\em accessibility relation} which determines the nature of the modal
logic. Each world is given a unique label, taken from a set $S$, which is usually
countably infinite. The accessibility relation $R$ is a binary relation on $S$.
The pairing of $S$ and $R$ defines a {\em frame} or structure which underpins
the model of modal logic. To complete the model we add an interpretation
\[h: S \times PROP \rightarrow \{\textrm{true}, \textrm{false}\} \]
of propositional formulae $\in PROP$ in each state. 
\\
Given $s \in S$ and 
$a \in PROP$, 
\begin{tabbing}
Spaziolo \= Formulalalalalala \= *iff* \=  \kill 
\> $\langle S, R, h \rangle \models_s a$ \> iff \> $h(s,a) = true$
\end{tabbing}  
This is read as: $a$ is true in world $s$ in the model $\langle S, R, h \rangle$ 
iff $h$ maps $a$ to true in world $s$. In general when a formula $\varphi$ is
true in a world $s$ in a model $\cal M$, it is denoted by
\[ {\cal M} \models_s \varphi \]
and if it is true in every world in the set $S$, it is said to be true in the
model, and denoted by
\[ {\cal M} \models \varphi \]
The boolean connectives are given the usual meaning:
\begin{tabbing}
Spaziolo \= Formulalalalalalalalalo \= *iff* \=  \kill 
\> $ \langle S, R, h \rangle \models_s \varphi \vee \psi $ \>
iff \> $ \langle S, R, h \rangle \models_s \varphi
~~\textrm{or}~~\langle S, R, h \rangle \models_s \psi $
\\
\> $ \langle S, R, h \rangle \models_s \varphi \Rightarrow \psi $ \>
iff \> $ \langle S, R, h \rangle \models_s \varphi
~~\textrm{implies}~~\langle S, R, h \rangle \models_s \psi $ 
\end{tabbing}
The frame enters the semantic definition only when the modality \nec\ is used,
as the formula \nec $\varphi$ is true in a world $s$ exactly when every world
$t$ in $S$ which is accessible from $s$ (i.e. such that $s~R~t$) has $\varphi$
true. More formally,
\begin{tabbing}
Spaziolo \= Formulalalalalalala \= *iff* \=  \kill
\> $ \langle S, R, h \rangle \models_s \Box \varphi $ \>
iff \> for all $~~t \in S, ~~~~s~R~t~~\textrm{implies}~~
\langle S, R, h \rangle \models_t \varphi $
\end{tabbing}
The models $ \langle S, R, h \rangle$ and the semantics we introduced for connectives 
are also known as Kripke models (or structures) and Kripke semantics, respectively
\cite{kripke63a,kripke63b,kripke65}, from the name of the author who mainly
contributed to developing a satisfactory  semantic theory of modal logic.

\subsection{\agento}
\label{agento}

Shoham's paper {\em Agent-Oriented Programming} \cite{shoham93}
is one of the most cited papers in the agent community, since
it proposed a new programming paradigm that 
\begin{quote}
promotes a
societal view of computation, in which multiple ``agents''
interact with one another.
\end{quote}  
In this section we first
introduce the basic concepts of the agent-oriented programming
(AOP) paradigm, and then we present the \agento\
programming language.  This is often referred to as the first
agent programming language, even though 
\begin{quote}
the simplifications
embodied in \agento\ are so extreme that it may be
tempting to dismiss it as uninteresting \cite{shoham93}.
\end{quote}

We opted for describing \agento\ as the language representing  
the class of languages based on mental modalities because it was
the first one to adopt this approach.
Other agent programming languages including mental modalities
are \textsf{3APL} 
\cite{ST41} and \textsf{AgentSpeak(L)} \cite{agentspeakl} which will
be discussed in Section \ref{related-and-future}.


For Shoham, a complete AOP system will include three primary components:
\begin{enumerate}
\item A restricted formal language with clear syntax and semantics for describing 
mental states; the mental state will be defined uniquely by several modalities, 
such as belief and commitments.
\item An interpreted programming language in which to define and program agents,
 with primitive commands such as \emph{REQUEST} and \emph{INFORM}.
\item An ``agentification process'' to treat 
existing hardware devices or software applications like agents. 
\end{enumerate}
The focus of Shoham's work is on the second component.

The mental categories upon which the AOP is based are {\em belief} and 
{\em obligation} (or {\em commitment}). A third category, which is not a mental 
construct, is {\em capability}. {\em Decision} (or {\em choice}) is treated as 
obligation to oneself.

Since {\em time} is basic to the mental categories, it is necessary to specify it. 
A simple point-based temporal language is used to talk about time; a typical sentence 
will be

\[ holding( robot, cup)^t \]

meaning that the robot is holding the cup at time $t$.

As far as {\em actions} are concerned, they are not distinguished from facts: the 
occurrence of an action is represented by the corresponding fact being true.

{\em Beliefs} are represented by means of the modal operator $B$. The general form 
of a belief statement is

\[ B_a^t \varphi  \]

meaning that agent $a$ believes $\varphi$ at time $t$. $\varphi$ may be a sentence
like $holding( robot, cup)^t$ or a belief statement: nested belief statements
like $B_a^3 B_b^{10} like(a, b)^7$, meaning that at time 3 agent $a$ believes that
at time 10 agent $b$ will believe that at time 7 $a$ liked $b$, are perfectly legal
in the AOP language.

The fact that at time $t$  an agent $a$  commits himself to agent $b$ about $\varphi$
is represented by the sentence

\[ OBL_{a, b}^t \varphi \]

A {\em decision} is an obligation to oneself, thus

\[ DEC_a^t f  \stackrel{def}{=}  OBL_{a,a}^t f   \]

The fact that at time $t$ agent $a$ is {\em capable} of $\varphi$ is represented by

\[ CAN_a^t \varphi \]

Finally, there is an ``immediate'' version of $CAN$: 

\[ ABLE_a \varphi  \stackrel{def}{=}  CAN_a^{time(\varphi)} \varphi    \]

where $time(B_a^t\psi) = t$ and $time(pred(arg_1, ..., arg_n)^t) = t$).

To allow the modalities introduced so far resemble their common sense counterparts, 
some assumptions are made:
\begin{itemize}
\item {\em Internal consistency}: both the beliefs and the obligations are assumed to 
be internally consistent.
\item {\em Good faith}: agents commit only to what they believe themselves capable of, 
and only if they really mean it.
\item {\em Introspection}: agents are aware of their obligations.
\item{\em Persistence of mental state}: agents have perfect memory of, and faith in, 
their beliefs, and only let go off a belief if they learn a contradictory fact. 
Obligations too should persist, and capabilities too tend not to fluctuate wildly.
\end{itemize}


\agento\ is a simple programming language that implements some of the
AOP concepts described above. Since \agento\ allows to define one program
for each agent involved in the system, it is no longer necessary to 
explicitly say which agent is performing which 
action; as an example, the statement $B_a^t \varphi$ becomes
$(B (t (\varphi)))$ in the body of code associated with agent $a$. 
In \agento\ the programmer specifies only conditions for making commitments; 
commitments are actually made and later carried out, automatically at the appropriate 
times. Commitments are only to primitive actions, those that the agent can directly 
execute. 
Before defining the syntax of commitments, other basic definitions
 are necessary.

\begin{description}
\item[Facts.] Fact statements constitute a tiny fragment of the temporal language 
described in the previous paragraph: they are essentially the atomic objective sentences
of the form \\
\centerline{\emph{(t atom)}}
and \\
\centerline{\emph{(NOT (t atom))}} 
For example, {\em (0 (stored orange 1000))} is an \agento\ fact stating that at time
0 there where 1000 oranges in the warehouse.

\item[Private actions.] The syntax for private actions is
\\ 
\centerline{\emph{(DO t p-action)}}
where \emph{t} is a time point and \emph{p-action} is a private action name.
 The effects of private actions may or may not be visible to other agents.

\item[Communicative actions.] There are three types of communicative actions:
\\
\centerline{\emph{(INFORM t a fact)}}
where \emph{t} is the time point in which informing takes place, \emph{a}
 is the receiver's name and \emph{fact} is a fact statement. 
\\              
\centerline{\emph{(REQUEST t a action)}}
where \emph{t} is a time point, \emph{a} is the receiver's name and 
\emph{action} is an action
 statement. 
\\
\centerline{\emph{(UNREQUEST t a action)}    }
where \emph{t} is a time point, \emph{a} is the receiver's name and 
\emph{action} is an action
 statement. 

\item[Nonaction.]
A ``nonaction'' prevents an agent from committing to a particular action.
\\
\centerline{\emph{(REFRAIN action)} }

\item[Mental conditions.] A mental condition is a logical combination of 
{\em mental patterns} which  may assume two forms:
\\
\centerline{\emph{(B fact)}  }
 meaning that the agent believes \emph{B} or 
\\
\centerline{\emph{((CMT a) action)}}
 where \emph{CMT} stands for commitment. The information about time
is included in facts and actions; an example of a mental pattern is
\emph{(B (3 (stored orange 950)))} meaning that the agent believes that at 
time 3 there were 950 oranges left in the warehouse.

\item[Capabilities.] The syntax of a capability is \\
\centerline{\emph{(action mentalcondition)}}
meaning that the agent is able to perform \emph{action} 
provided that \emph{mentalcondition} (see above) is true. Throughout the 
example we will use this
syntax which is inherited from the original paper, even if a notation including
the {\em CAN} keyword (namely, \emph{(CAN action mentalcondition)})
would be more appropriate. 
 
\item[Conditional action.]
The syntax of a conditional action is
\\
\centerline{\emph{(IF mentalcondition action)} }
meaning that \emph{action} can be performed only if 
\emph{mentalcondition} holds.
\item[Message condition.] A message condition is a logical 
combination of {\em message patterns}, which are triples
\\
\centerline{\emph{(From Type Content)} }
where \emph{From} is the sender's name, \emph{Type} is 
\emph{INFORM, REQUEST} or \emph{UNREQUEST} and 
\emph{Content} is a fact statement or an action statement. 

\item[Commitment rule.]
A commitment rule has the form:
\\
\centerline{\emph{(COMMIT messagecondition mentalcondition (agent action)*)}}
where \emph{messagecondition} and \emph{mentalcondition} are respectively 
message and mental conditions, 
\emph{agent} is an agent name, \emph{action}
 is an action statement and \emph{*} denotes repetition of zero or more times.
The intuition behind the commitment rule 
\emph{(COMMIT msgcond mntcond (ag$_1$ act$_1$) ... (ag$_n$ act$_n$))} in the program defining 
the behavior of agent {\em ag} is that if {\em ag}
receives a message satisfying {\em msgcond} and its mental states 
verifies the condition {\em mntcond}, it commits to agent {\em ag$_1$}
about {\em act$_1$}, ..., and to agent {\em ag$_n$} about {\em act$_n$}.
Note that commitment rules are identified by the {\em COMMIT} keyword 
and commitment mental pattern (see the definition of mental conditions above) 
are identified by the {\em CMT} keyword. We adopt
this syntax to be consistent with the original paper even if we are aware that 
using similar keywords for different syntactic objects may be confusing.

\item[Program.]
A program is defined by the time unit, called  ``timegrain'', followed by 
the capabilities, the initial beliefs and 
the commitment rules of an agent. Timegrain ranges over {\em m} (minute), {\em h} (hour), 
{\em d} (day) and {\em y} (year).
\end{description}

\subsubsection{Semantics}

No formal semantics for the language is given.

\subsubsection{Implementation}

A prototype \agento\ interpreter has been implemented in \textsf{Common Lisp}
and has been installed on Sun/Unix, DecStation/Ultrix and Macintosh computers.
Both the interpreter and the programming manual are available to the scientific community.
A separate implementation has been developed by Hewlett Packard as part of a joint 
project to incorporate \textsf{AOP} in the New Wave$^{TM}$ architecture.

The \agento\ engine is characterized by the following two-step
cycle:
\begin{enumerate}
\item Read the current messages and update beliefs and commitments.
\item Execute the commitments for the current time, possibly resulting in further 
belief change.
\end{enumerate}
Actions to which agents can be committed include communicative ones such as informing and
requesting, as well as arbitrary private actions.

\subsubsection{Extensions}

Two extensions of \agento\ have been proposed:
\begin{itemize}
\item 
\textsf{PLACA} \cite{placa} enriches \agento\ with a mechanism for
flexible management of plans. It adds two data structures to
 the agent's state: a list of intentions and a list of  
plans. Intentions are adopted is a similar way that commitments are; the
\textsf{PLACA} command 
{\em (ADOPT (INTEND x))} means that the agent will add the intention to do
{\em x} to its intention list. Plans are created by an external plan generator 
to meet these intentions. This approach gives the system the ability to dynamically 
alter plans that are not succeeding.
\item \textsf{Agent-K} \cite{agentk} is an attempt to standardize the message 
passing functionality in \agento. It combines the syntax of \agento\ 
(without support for the planning mechanisms of \textsf{PLACA}) 
with the format of \kqml\ ({\em Knowledge Query and Manipulation Language} \cite{Mayfield:95})
to ensure that messages written in 
languages different from \agento\ can be handled.
\textsf{Agent-K} introduces two major changes to the structure of \agento: first,
it replaces outgoing {\em INFORM}, {\em REQUEST}, and {\em UNREQUEST} message actions 
with one command, {\em KQML}, that takes as its parameters the message, 
the time, and the {\em KQML} type; second,  
it allows many commitments to match a single message. 
In \agento\ the multiple commitment mechanism 
was not defined and the interpreter simply 
selected the first rule that matched a message.

\end{itemize}

\subsubsection{Example}

The \agento\ program for the seller agent may be as follows.
Variables are preceded by a ``?'' mark instead of being uppercase, coherently
with the language syntax. Universally quantified variables, whose scope is 
the entire formula,  
are denoted by the prefix ``?!''. 
\\
\\
{\em
timegrain    := m
}
\\
\\
The program is characterized by a 
time-grain of one minute.
\\
\\
\\
{\em 
CAPABILITIES := ((DO ?time (ship ?!buyer ?merchandise ?required-amount ?!price)) \\
\hspace*{1.2cm}                 (AND (B (?time (stored ?merchandise ?stored-amount))) \\
\hspace*{3.5cm}                        ($>=$ ?stored-amount ?required-amount)))}
\\
\\
The agent has the capability of shipping a certain amount of 
merchandise, provided that, at the time of shipping, it believes that 
such amount is stored in the warehouse. 
\\
\\
\\
{\em INITIAL BELIEFS := (0 (stored orange 1000))\\
\hspace*{1.2cm}                  (?!time (min-price orange 1))\\
\hspace*{1.2cm}                  (?!time (max-price orange 2))}
\\
\\
The agent has 
initial beliefs about the minimum price, maximum price and stored amount of
oranges. The initial belief about stored oranges only holds at time 0, since
this amount will change during the agent's life, while beliefs about minimum and 
maximum prices hold whatever the time.
\\
\\
\\
{\em COMMITMENT RULES := (COMMIT \\
\hspace*{1.2cm}                     (?buyer REQUEST \\
\hspace*{2.7cm}                       (DO now (ship ?buyer ?merchandise ?req-amnt ?price)))
\\
\\
\hspace*{1.2cm}                     (AND (B (now (stored ?merchandise ?stored-amount))) \\
\hspace*{3.5cm}                          ($>=$ ?stored-amount ?req-amnt)\\
\hspace*{3.5cm}                          (B (now (max-price ?merchandise ?max)))\\
\hspace*{3.5cm}                          ($>=$ ?price ?max))
\\
\\
\hspace*{1.2cm}                     (?buyer (DO now \\
\hspace*{2.7cm}                        (ship ?buyer ?merchandise ?req-amnt ?price)))\\
\hspace*{1.2cm}                     (myself (INFORM now ?buyer\\ 
\hspace*{2.7cm}                        (accepted ?merchandise ?req-amnt ?price)))\\
\hspace*{1.2cm}                     (myself (DO now (update-merchandise ?merchandise ?req-amnt)))\\
\hspace*{1.2cm}                  )}
\\
\\
The first commitment rule says that if the seller agent 
receives a request of shipping a certain amount of merchandise at a certain price,
 and if it believes that the required 
amount is stored in the warehouse and the proposed 
price is greater than {\em max-price}, the seller agent commits
itself to the buyer to 
ship the merchandise (\emph{ship} is a private action), and decides
(namely, commits to itself) to 
 inform the buyer that its 
request has been accepted and to
update the stored amount of merchandise (\emph{update-merchandise} is a private action).
\\
\\
\\
{\em 
\hspace*{1.2cm}                  (COMMIT \\
\hspace*{1.2cm}                    (?buyer REQUEST \\
\hspace*{2.7cm}                        (DO now (ship ?buyer ?merchandise ?req-amnt ?price)))\\
\\
\hspace*{1.2cm}                     (OR (AND (B (now (stored ?merchandise ?stored-amount))) \\
\hspace*{3.5cm}                              ($<$ ?stored-amount ?req-amnt))\\
\hspace*{2.7cm}                         (AND (B (now (min-price ?merchandise ?min)))\\
\hspace*{3.5cm}                              ($<=$ ?price ?min)))
\\
\\
\hspace*{1.2cm}                     (myself (INFORM now ?buyer \\
\hspace*{2.7cm}                        (refused ?merchandise ?req-amnt ?price)))\\
\hspace*{1.2cm}                  )}
\\
\\
The second rule says that if the required amount of merchandise is not present in the 
warehouse or the price is too low, the seller agent decides to inform the
buyer that its request has been refused.
\\
\\
\\
{\em \hspace*{1.2cm}                  (COMMIT \\
\hspace*{1.2cm}                     (?buyer REQUEST \\
\hspace*{2.7cm}                        (DO now (ship ?buyer ?merchandise ?req-amnt ?price)))\\
\\
\hspace*{1.2cm}                     (AND (B (now (stored ?merchandise ?stored-amount))) \\
\hspace*{3.5cm}                          ($>=$ ?stored-amount ?req-amnt)\\
\hspace*{3.5cm}                          (B (now (max-price ?merchandise ?max)))\\
\hspace*{3.5cm}                          ($<$ ?price ?max)\\
\hspace*{3.5cm}                          (B (now (min-price ?merchandise ?min)))\\
\hspace*{3.5cm}                          ($>$ ?price ?min))\\
\\                          
\hspace*{1.2cm}                     (myself (DO now (eval-mean ?max ?price ?mean-price)))\\
\hspace*{1.2cm}                     (myself (REQUEST now ?buyer \\
\hspace*{2.7cm}                        (eval-counter-proposal ?merchandise ?req-amnt ?mean-price)))\\
\hspace*{1.2cm}                    )}
\\
\\
Finally, the third rule
says that if the price can be negotiated and
there is enough  merchandise in the warehouse, the seller agent evaluates the
price to propose to the buyer agent (\emph{eval-mean} is a private action)
 and decides to send a 
counter-proposal to it. We are assuming that the buyer agent is able to 
perform an {\em eval-counter-proposal} action to evaluate the 
seller agent's proposal: the seller agent must know the exact action syntax
if it wants that the buyer agent understands and satisfies its request.


\section{Deontic Logic}
\label{deontic-logic}

This introduction is based on the book {\em Deontic logic in
Computer Science} \cite{deonticBook}.
Deontic logic is the logic to reason about ideal and actual behavior. From the
1950s, von Wright \cite{vonWright51}, Casta\~neda
\cite{castaneda75}, Alchourr\'on and Bulygin \cite{alchourron71} and others developed
deontic logic as a modal logic with operators for permission, obligation and
prohibition. Other operators are possible, such as formalizations of the system
of concepts introduced by Hohfeld in 1913, containing operators for duty, right,
power, liability, etc \cite{hohfeld13}.
Deontic logic has traditionally been used to analyze the structure of normative
law and normative reasoning in law. Recently it has been realized that deontic
logic can be of use outside the area of legal analysis and legal automation:
it has a potential use in any area where we want to reason about ideal as well
as actual behavior of systems. 
To give an idea of what deontic logic systems look like, 
we describe the OS Old System  \cite{vonWright51} and the
KD Standard
System of Deontic Logic \cite{aqvist84}.

\subsection{The OS System} 

The OS system is based on two deontic operators: $\bfO$, meaning
obligation, and $\bfP$ meaning permission. Let $p$ be a proposition in the
propositional calculus, then 
$\bfO p$ and
$\bfP p$
are formulae in the OS deontic logic language.  

The system consists of the following
axioms and inference rule: 
\begin{description}
\item[(OS0)] All tautologies of Propositional Calculus
\item[(OS1)] $\bfO p \equiv \neg\bfP \neg p$
\item[(OS2)] $\bfP p  \vee \bfP \neg p$
\item[(OS3)] $\bfP(p \vee q) \equiv \bfP p \vee \bfP q$
\item[(OS4)] \large{$\frac{p \equiv q}{\bfP p \equiv \bfP q}$} 
\end{description}
Axiom (OS1) expresses that having an obligation to $p$ is equivalent to not
being permitted to not $p$; (OS2) states that either $p$ is permitted or not $p$
is; (OS3) says that a permission to $p$ or $q$ is equivalent to $p$ being
permitted or $q$ being permitted; (OS4) asserts that if two
assertions are equivalent, then permission for one implies permission for the
other, and vice versa. 
Later, it was realized that the system OS is very close to a normal modal logic,
enabling a clear Kripke-style semantics using $\bfO$ as the basic necessity
operator, at the expense of introducing the validity of 
\begin{description}
\item[(O$\top$)] $\bfO (p \vee \neg p)$
\end{description}
stating the existence of an empty normative system, which von Wright rejected as
an axiom. 

\subsection{The KD System}
The KD System is a von Wright-type system including the $\bfF$ (forbidden) operator  
and consisting of the following axioms and
rules.
\begin{description}
\item[(KD0)] All tautologies of Propositional Calculus
\item[(KD1)] $\bfO(p \Rightarrow q) \Rightarrow (\bfO p \Rightarrow \bfO q)$
\item[(KD2)] $\bfO p \Rightarrow \bfP p$
\item[(KD3)] $\bfP p \equiv \neg \bfO \neg p$
\item[(KD4)] $\bfF p \equiv \neg \bfP p $
\item[(KD5)] Modus ponens: {\large $\frac{p~~~~~~p \Rightarrow q}{q}$} 
\item[(KD6)] O-necessitation: {\large $\frac{p}{\bfO p}$} 
\end{description} 
Axiom (KD1) is the so called {\em K-axiom};  (KD2) is the {\em
D-axiom}, stating that obligatory implies permitted; (KD3) states that
permission is the dual of obligation and (KD4) says that forbidden is not
permitted. (KD1) holds for any modal necessity operator. Essentially, it states
that obligation is closed under implication. Whether this is desirable may be
debatable, but it is a necessary consequence of the modal approach. Note
furthermore that the O-necessitation rule (KD6), which is also part of the 
idea of viewing deontic logic as a normal modal logic, implies the axiom
rejected by von Wright
\begin{description}
\item[(O$\top$)] $\bfO (p \vee \neg p)$
\end{description}
So, if we want to view deontic logic as a branch of Kripke-style modal logic, we
have to commit ourselves to (O$\top$).

As with other modal logics, the semantics of the standard system is based on the
notion of a possible world. Given a Kripke model 
$ \langle S, R, h \rangle$ and a world $s \in S$ we give the following semantics
to the modal operators:  
\begin{tabbing}
Spaziolo \= Formulalalalalalalala \= *iff* \=  \kill 
\> 
$ \langle S, R, h \rangle \models_s \bfO p $ \>
iff \> 
$\textrm{for all}~~t \in S,~~s ~R ~t~~\textrm{implies}~~  
\langle S, R, h \rangle \models_t p$
\\
\> 
$ \langle S, R, h \rangle \models_s \bfP p $ \>
iff \> 
$ \textrm{exists}~~t \in S ~~\textrm{such that}~~ s ~ R~t~~ \wedge 
~~ \langle S, R, h \rangle \models_t p$
\\
\> 
$ \langle S, R, h \rangle \models_s \bfF p $ \>
iff \> $
 \textrm{for all}~~t \in S,~~s~ R~ t~~\textrm{implies}~~  
 \langle S, R, h \rangle
\not\models_t p$
\end{tabbing}

As already stated, the operator $\bfO$ is treated as the basic modal operator $\Box$:
for $\bfO p$ being true in world $s$ we have to check whether $p$ holds in all
the worlds reachable from $s$, as given by the relation $R$. This reflects the
idea that something is obligated if it holds in all perfect (ideal) worlds
(relative to the world where one is). This semantics 
is exactly the same semantics of $\Box$, as defined in Section \ref{modal-logic}.
 The other operators are more or less
derived from $\bfO$. The operator $\bfP$ is the dual of $\bfO$: $\bfP p$ is true
in the world $s$ if there is some world reachable from $s$ where $p$ holds.
Finally, something is forbidden in a world $s$ if it does not hold in any world
reachable from $s$.

\subsection{The \impact\ Agent Language}
\label{impact}

We introduce the \impact\ agent programming language 
\cite{aris-etal-99,eite-etal-98o,eite-subr-99,eite-etal-00a} 
as a relevant example of 
 use of deontic logic to specify agents. 
In order to describe this language, we provide a set of definitions on top of which the 
language is based.

\comment{
\paragraph*{The \impact\ phylosophy.}

As different application programs reason with different types of data, 
and even programs dealing with the same types of data often
manipulate them in a variety of ways, it is critical that any
notion of agenthood be applicable to arbitrary software programs.
Agent developers should be able to select data structures that best
suit the application functions desired by users of the application
they are building.  Figure \ref{impactarch} shows the architecture of a
full-fledged \impact\ software agent.  It is important to note that all agents
have the same architecture and hence the same components, but the {\em
content} of these components can be different, leading to different
behaviors and capabilities offered by different agents.

\begin{figure}
\centerline{\psfig{file=./Figs/f2-2.ps,width=12cm}}
\caption{Basic Architecture of \impact\ Agents}
\label{impactarch}
\end{figure}
}

\paragraph*{Agent Data Structures.}
All \impact\ agents are built ``on top'' of some existing body of code
specified by 
 the data types or data structures,
$\sctype$, that the agent manipulates and by a set of
functions, $\scfunct$, that are callable by external programs. Such functions
constitute the {\em application programmer interface} or API of the
package on top of which the agent is being built.

Based on $\sctype$ and $\scfunct$ supported by a package (a body of
software code) $\scode$, we may use a unified language to query
the data structures.  If $\tt f \in \scfunct$ is an $n$-ary function defined in
that
package, and $\tt t_1$, \ldots, $\tt t_n$ are {\em terms}
 of appropriate types, then $\tt
\scode:f(t_1,\ldots ,t_n)$
is a {\em code call}.  This code call says ``Execute function $\tt f$
as defined in package $\scode$ on the stated list of arguments.'' 

A {\em code call atom} is an expression $\cca$ of the form ${\tt
in(t,\ccall)}$ or ${\tt notin(t,\ccall)}$, where ${\tt t}$ is a term and
$\ccall$ is a code call. ${\tt in(t,\ccall)}$ evaluates to true (resp. false) in a given 
state
if ${\tt t}$ is (resp. is not) among the values returned by calling ${\tt \ccall}$
in that state. The converse holds for ${\tt notin(t,\ccall)}$.
For example,
\\
\centerline{${\tt in(\langle InOut, Sender, Receiver, Message, Time\rangle, msgbox:getMessage(Sender))}$}
\\
is true in a given state if the term ${\tt \langle InOut, Sender, Receiver, Message, Time \rangle}$ is among 
the values returned by calling the ${\tt getMessage(Sender)}$ function 
provided by the ${\tt msgbox}$ package in that state.

A \emph{code call condition} is a conjunction of code call atoms and
{\em constraint atoms} of
the form ${\tt t_1}~op~{\tt t_2}$ where $op$ is any of $=,~\neq,~<,~\leq,~>,~\geq$
and ${\tt t_1},{\tt t_2}$ are terms.  

Each agent is also assumed to have access to a message box
package identified by ${\tt msgbox}$, together with some API function calls to
access it (such as the ${\tt getMessage}$ function appearing in the
code call atom above). Details of the message box in \impact\ may be found 
in \cite{eite-etal-98o}.

At any given point in time, the actual set of objects in the data
structures (and message box) managed by the agent
constitutes the {\em state} of the agent. We shall identify a state $\agstate$
with the set of ground (namely, containing no variables) 
code calls which are true in it.

\paragraph*{Actions.}  The agent can execute a set of {\em actions}
$\alpha(X_1,\ldots,X_n)$.  Such actions may include
reading a message from the message box, responding to a message,
executing a request, updating the agent data structures, etc.  Even doing
nothing may be an action. Expressions $\alpha(\vect)$, where $\vect$ is a
list of terms of appropriate types, are {\em action atoms}. 
 Every action $\alpha$ has a precondition
$\pre{\alpha}$ (which is a code call condition), a set of effects
(given by an add list $\add{\alpha}$ and a delete list $\del{\alpha}$
of code call atoms) that describe how the agent state changes when the
action is executed, and an {\em execution script or method} consisting
of a body of physical code that implements the action.

\paragraph*{Notion of Concurrency.}  The agent has an associated
body of code implementing a {\em notion of concurrency}
$\concur(AS,\agstate)$. Intuitively, it takes a set of actions $AS$
and the current agent state $\agstate$ as input, and returns a single
action (which ``combines'' the input actions together) as output.
Various possible notions of concurrency are described in
\cite{eite-etal-98o}. 
For example, weak concurrent execution is defined as follows. Let
$AS$ be the set of actions in the current status set, evaluated according 
to the chosen semantics. Weakly concurrently executing actions in 
$AS$ means that first all the deletions in the delete list of actions in 
$AS$ are done in parallel
and then all the insertions in the add list of actions in $AS$ are. Even if
some problems arise with this kind of concurrency, it has the advantage that deciding
whether a set of actions is weakly-concurrent executable is polynomial
(Theorem 3.1 of \cite{eite-etal-98o}).

\paragraph*{Integrity Constraints.}
Each agent has a finite set $\IC$ of {\em integrity constraints}  that
the state $\agstate$ of the agent must satisfy (written  $\agstate\models \IC$), of the form
$
\psi \; \Rightarrow \; \chi_a
$
where $\psi$ is a code call condition, and $\chi_a$ is a code call
atom or constraint atom. Informally, $\psi \; \Rightarrow \; \chi_a$ 
has the meaning of the
universal statement ``If $\psi$ is true, then $\chi_a$ must be
true.'' For simplicity, we omit here and in other places safety
aspects (see 
\cite{eite-etal-98o} for details). 

\paragraph*{Agent Program.}  Each agent has a set of rules called
the {\em agent program} specifying the principles under which the
agent is operating.  These rules specify, using deontic modalities,
what the agent may do, must do, may not do, etc.  Expressions
$\bfO\alpha(\vect)$, $\bfP\alpha(\vect)$,
$\bfF\alpha(\vect)$, $\bfDo\alpha(\vect)$, and $\bfW\alpha(\vect)$,
where $\alpha(\vect)$ is an action atom, are
called \emph{action status atoms}.  These action status atoms are read
(respectively) as $\alpha(\vect)$ is \emph{obligatory, permitted,
forbidden, done}, and the obligation to do $\alpha(\vect)$ is
\emph{waived}.  If $A$ is an action status atom, then $A$ and $\neg A$
are called \emph{action status literals}.  An \emph{agent program}
$\agprog$ is a finite set of rules of the form:
\begin{eqnarray}
\label{rule}
A & \leftarrow & \chi\,\&\, L_1\,\&\, \cdots \,\&\, L_n
\end{eqnarray}
where $A$ is an action status atom, $\chi$ is a code call condition,
and $L_1,\ldots ,L_n$ are action status literals.

\subsubsection{Semantics}
\label{impactSemantics}

If an agent's behavior is defined by a program $\agprog$, the question 
that the agent must answer, over
and over again is:
\begin{quote}
{\em 
What is the set of all action status atoms of the form  $\bfDo\alpha(\vect)$
that are true with respect to $\agprog$, the current state $\agstate$
 and the set $\IC$ of underlying integrity constraints
on agent states?
}
\end{quote}
This set defines the actions the agent must take; 
\cite{eite-etal-98o} provides a series of successively more refined semantics
for action programs that answer this question, that we 
discuss in a very  succinct form.  

\begin{definition}[Status Set]
{\rm
A {\em status set}\/ is any set $S$ of
ground action status atoms over the values from the type domains of
a software package $\SC$.  
}
\end{definition}

\begin{definition}[Operator \App]
{\rm 
Given a status set $S$, the operator \App\ computes all
action status atoms that may be inferred to be true by allowing the 
rules in \agprog\ to fire exactly once. It is defined in the following way:
let $\agprog$ be an agent program and $\agstate$ be an agent
state. Then, $\App = \{ Head(r\theta)$
 $\mid\; r \in \agprog,\ R(r,\theta,S)
\textrm{ is true
on } \agstate \}$,
where $Head(A \leftarrow 
 \chi\,\&\, L_1\,\&\, \cdots \,\&\, L_n) = A$ and the predicate $R(r,\theta,S)$ is true iff
(1) $r\theta: A \la \chi \,\&\, L_1\,\&\,\cdots \,\&\, L_n$ is a ground rule,
(2) $\agstate\models \chi$,
(3) if $L_i=Op(\alpha)$ then $Op(\alpha) \in S$, and (4) if $L_i=\neg
Op(\alpha)$ then $Op(\alpha) \notin S$, for all $i \in \{1,\ldots,n\}$.
}
\end{definition}

\begin{definition}[\acl{S}{}]
\label{defn:acl}
{\rm 
A status set $S$ is {\em deontically closed}, if for every ground action
$\alpha$,
it is the case that $(DC1)$ $\bfO\alpha\in S$ implies $\bfP\alpha\in S$.
A status set $S$ is {\em action closed}, if for every ground action
$\alpha$, it is the case that
$(AC1)$ $\bfO\alpha\in S$ implies $\bfDo\alpha\in S$, and
$(AC2)$ $\bfDo\alpha\in S$ implies $\bfP\alpha\in S$.
It is easy to notice that status sets that are action closed are also
deontically closed.
For any status set $S$, we denote by $\acl{S}{}$ the smallest set
$S'\supseteq S$ such that $S'$ is closed under $(AC1)$ and $(AC2)$,
i.e., {\em action closed}.
}
\end{definition}

\begin{definition}[Feasible Status Set]
\label{defn:feasible}
{\rm
Let $\agprog$ be an agent program and let $\Ob$ be an agent
state. Then, a status set $S$ is a {\em feasible status set} for $\agprog$
on $\Ob$, if $(S1)$-$(S4)$ hold:

\begin{description}
 \item[{\rm $(S1)$}]\ $\App \subseteq S$;

    \item[{\rm $(S2)$}] \ For any ground action $\alpha$, the following holds:
    $\bfO\alpha \in S$
implies $\bfW\alpha \notin S$, and $\bfP\alpha \in S$ implies $\bfF\alpha \notin
S$.

\item[{\rm $(S3)$}] \ $S = \acl{S}{}$, i.e., $S$
is action closed;

 \item[{\rm $(S4)$}] The state $\Ob' =\concur(\bfDo(S),\OS)$ which results from
 $\agstate$ after
executing (according to some execution strategy $\concur$) the actions
in $\{\alpha \mid \bfDo(\alpha) \in S\}$ satisfies the integrity constraints, i.e.,
$\Ob'\models \intcons$. 
\end{description}
}
\end{definition}

\begin{definition}[Groundedness; Rational Status Set]
{\rm
A status set $S$ is {\em grounded}, if no
status set $S'\neq S$ exists such that $S' \subseteq S$ and $S'$
satisfies conditions $(S1)$--$(S3)$ of a feasible status set.
A status set $S$ is a {\em rational status set}, if
$S$ is a feasible status set and $S$ is grounded.
}
\end{definition}

\begin{definition}[Reasonable Status Set]
\label{RSS}
{\rm
Let $\agprog$ be an agent program, let $\agstate$ be an agent state, and
let
$S$ be a status set.\\
\\
\noindent 1. If $\agprog$ is positive, i.e., no negated
action status atoms occur in it, then $S$ is a {\em reasonable
status set} for $\agprog$ on $\agstate$, iff $S$ is
a rational status set for $\agprog$ on $\agstate$.
\\
\noindent 2. The reduct of $\agprog$ w.r.t.\ $S$ and $\agstate$, denoted by
$red^S(\agprog,\agstate)$, is the program which is obtained from the
ground instances of the rules in $\agprog$ over $\agstate$ as follows.
\\
\noindent \hspace*{.5cm}(a)  Remove every rule $r$ such that $Op(\alpha) \in S$ for some
    $\neg
        Op(\alpha)$ in the body of $r$;\\
\noindent \hspace*{.5cm}(b) remove all negative literals $\neg Op(\alpha)$ from the remaining
        rules.\\
Then $S$ is a {\em reasonable status set} for $\agprog$ w.r.t.\ $\agstate
$,
if it is a reasonable status set of the program $red^S(\agprog,\agstate)$
with respect to $\agstate$. 
}
\end{definition}

\subsubsection{Implementation}

The implementation of the \impact\ agent program consists of two major parts,
both implemented in \java:
\begin{enumerate}
\item the \impact\ Agent Development Environment (IADE for short) which is used by the 
developer to build and compile agents, and
\item the run-time part that  allows the agent to autonomously update its reasonable status
set and execute actions as its state changes.
\end{enumerate} 

The IADE provides a network accessible interface through which an agent developer can 
specify the data types, functions, actions, integrity constraints, notion of concurrency 
and agent program associated with her/his agent; it also provides support for 
compilation and testing.

The runtime execution module runs as a background applet and performs the following
steps: (i) monitoring of the agent's message box, (ii) execution of the algorithm for
updating the reasonable status set and (iii) execution of the actions $\alpha$ such that
${\bf Do}\alpha$ is in the updated reasonable status set.

\subsubsection{Extensions}

Many extensions to the \impact\ framework are 
discussed in the book \cite{impactbook} which analyses:
\begin{itemize}
\item {\em meta agent programs} to reason about other agents 
based on the beliefs they hold;
\item {\em temporal agent programs}  to specify temporal aspects
of actions and states;
\item {\em probabilistic agent programs} to deal with uncertainty; and
\item {\em secure agent programs} to provide agents with security mechanisms.
\end{itemize}
Agents able to recover from an integrity constraints violation and able to 
continue to process some requests while continuing to recover are discussed 
in \cite{LNAI2407}. The integration of planning algorithms in the 
\impact\ framework is discussed in \cite{dixmunnau01}.

\subsubsection{Example}
The \impact\ example appears to be more complicated than the other ones because 
we exemplify how it is possible to specify actions that require an access 
to external packages. These actions are defined in terms of their
preconditions, add and delete list which involve the code call atoms that 
allow the real integration of external software. The ability
of accessing real software makes \impact\ specifications more complex than the
others we discuss in this paper but, clearly, also more powerful.  
We suppose that the \impact\ program for the seller agent 
accesses three software packages:
\begin{itemize}
\item an  {\em oracle} database where information on the stored amount
of merchandise and its minimum and maximum price is 
maintained in a {\em stored\_merchandise} relation;
\item a {\em msgbox} package that allows agents to exchange messages, as
described in Section 3 of \cite{eite-etal-98o};
in particular, it provides the {\em getMessage(Sender)} function
which allows all tuples coming from {\em Sender} to be read and 
deleted from the message box of the receiving agent; and
\item a mathematical package {\em math} providing mathematical functions. 
\end{itemize}

\begin{itemize}
\item Initial state:
\begin{itemize}
\item[] The {\em stored\_merchandise} relation, with schema 
$<${\em name, amount, min, max}$>$, initially contains the tuple 
$<${\em orange, 1000, 1, 2}$>$
\end{itemize}
\item Actions:
\begin{itemize}
	\item {\em ship(Buyer, Merchandise, Req\_amount)
\\
\\
Pre(ship(Buyer, Merchandise, Req\_amount)) = \\
\hspace*{.5cm}in(Old\_amount, \\
\hspace*{.5cm} oracle:select(stored\_merchandise.amount, name, =, Merchandise)) 
$\wedge$\\
\hspace*{.5cm}in(Difference, math:subtract(Old\_amount, Req\_amount)) $\wedge$\\
\hspace*{.5cm} Difference $\geq$ 0 \\
\\
Add(ship(Buyer, Merchandise, Req\_amount)) = \\
\hspace*{.5cm}in(Difference, \\
\hspace*{.5cm} oracle:select(stored\_merchandise.amount, name, =, Merchandise)) \\
\\
Del(ship(Buyer, Merchandise, Req\_amount)) = \\
\hspace*{.5cm}in(Old\_amount,\\
\hspace*{.5cm} oracle:select(stored\_merchandise.amount, name, =, Merchandise))
}
\\
In order for the agent to ship merchandise, there must be enough merchandise available: the precondition
of the action is true if the difference {\em Difference} between the stored amount
of merchandise, {\em Old\_amount}, and the required amount, {\em Req\_amount}, is
greater than or equal to zero. The effect of shipping is that the amount of available merchandise is
updated  by modifying the information in the {\em stored\_merchandise} table: 
{\em stored\_merchandise.amount} becomes equal to {\em Difference}, 
shared among the three equalities defining precondition, add list and delete list.
{\em Add} and {\em Del} denote the desired modifications to the database
through code calls. There is also a procedure, that we omit, that realizes this
action. In practice
this procedure would also issue an order to physically ship the merchandise.
\item[] ~~~
	\item {\em sendMessage(Sender, Receiver, Message)}\\ 
This action accesses the {\em msgbox} 
package putting a tuple in the agent 
message box. The {\em msgbox} package underlying this action is assumed
to ensure that tuples put in the message box are delivered to the
receiver agent. The precondition of this action is empty (it is always
possible to send a message), the add and delete lists consist of the 
updates to the receiver's mailbox.
\item[] The notion of concurrency we adopt is {\em weak concurrent execution} introduced in
Section \ref{impact}.
\end{itemize}

\item Integrity constraints:
\begin{itemize}
	\item[] 
{\em in(Min, oracle:select(stored\_merchandise.min, name, =, Merchandise)) $\wedge$ \\
in(Max, oracle:select(stored\_merchandise.max, name, =, Merchandise)) $\implies$ \\
\hspace*{.5cm} 0 $<$ Min $<$ Max
}
\\
This integrity constraint says that the minimum price allowed for any merchandise
must be greater than zero and lower than the maximum price. 
\item[] ~~~
\item[] {\em in(Amount, \\
\hspace*{.5cm} oracle:select(stored\_merchandise.amount, name, =, Merchandise)) $\implies$\\
\hspace*{.5cm} Amount $\geq$ 0
}
\\
This integrity constraint says that any amount of merchandise must be greater or 
equal to zero.
\item[]~~~
\item[] {\em in($\langle$o, Sender, Receiver, 
accept(Merchandise, Req\_amount, Price), T$\rangle$, \\
\hspace*{.5cm} msgbox:getMessage(Sender)) $\wedge$\\
in($\langle$o, Sender, Receiver, 
refuse(Merchandise, Req\_amount, Price), T$\rangle$, \\
\hspace*{.5cm} msgbox:getMessage(Sender) $\implies$\\
\hspace*{1.5cm} false
}
\\
This integrity constraint says that an agent cannot both accept and refuse an offer
(the {\em o} element in the tuple returned by the
{\em msgbox:getMessage(Sender)} code call means that the message is an output message from
{\em Sender} to {\em Receiver}; the last element of the tuple, {\em T}, is the time).
 Similar constraints could be added to enforce that
an agent cannot both accept and negotiate an offer 
and that it cannot both refuse and negotiate.
\item[]~~~
\item[] {\em  in(Min, oracle:select(stored\_merchandise.min, name, =, Merchandise)) $\wedge$\\
in($\langle$ o, Sender, Receiver, 
accept(Merchandise, Req\_amount, Price), T$\rangle$, \\
\hspace*{.5cm} msgbox:getMessage(Sender)) $\wedge$ Price $<$ Min $\implies$ \\
\hspace*{1.5cm} false}
\\
This integrity constraint says that an agent cannot accept
a proposal for a price lower than the minimum price
allowed. Different from the previous ones, this constraint involves two
different packages, the {\em oracle} one and the {\em msgbox} one. 
\item[]~~~
\item[] Other integrity constraints could be added to ensure the consistency of data
inside the same package or across different packages. Note that most of the above 
constraints are enforced by the agent program. The reason why they should be explicitly stated
is that, in the case of legacy systems, 
the legacy system's existing interface and the agent both access
and update the same data.  Thus, the legacy interface may alter
the agent's state in ways that the agent may find unacceptable. The
violation of the integrity constraits prevents the agent from
continuing its execution in an inconsistent state. 
\end{itemize}
\newpage
\item Agent Program:

\item[] ~~\\
\noindent {\em 
{\bf Do} 
sendMessage(Seller, Buyer, accept(Merchandise, Req\_amount, Price)) $\leftarrow$ \\
\hspace*{.5cm} in($\langle$i, Buyer, Seller, 
contractProposal(Merchandise, Req\_amount, Price), T$\rangle$,\\ 
\hspace*{1.5cm}msgbox:getMessage(Seller)), \\ 
\hspace*{.5cm} in(Max, oracle:select(stored\_merchandise.max, name, =, Merchandise)), \\
\hspace*{.5cm} in(Amount, oracle:select(stored\_merchandise.amount, name, =, Merchandise)), \\
\hspace*{.5cm} Price $>=$ Max, Amount $>=$ Req\_amount
}
\\
This rule says that if all the conditions for accepting a proposal are met,
namely
\begin{enumerate}
\item the seller agent
received a contractProposal from the buyer (the first element of 
the tuple in the first code call atom, {\em i}, says that the message is an input 
message),
\item there is 
enough merchandise in the warehouse and 
\item the proposed price is greater than the
{\em Max} value), 
\end{enumerate}
then the seller sends a message to the buyer, saying that it 
accepts the proposal.
\item[] ~~\\
\noindent {\em 
{\bf O} 
ship(Buyer, Merchandise, Req\_amount) $\leftarrow$ \\
\hspace*{.5cm} {\bf Do} sendMessage(Seller, Buyer, accept(Merchandise, Req\_amount, Price))
}
\\
This rule says that if the seller agent accepts the buyer's proposal
by sending a message to it, it is then obliged to 
ship the merchandise. 
\item[] ~~\\
\noindent {\em 
{\bf Do} 
sendMessage(Seller, Buyer, refuse(Merchandise, Req\_amount, Price)) $\leftarrow$ \\
\hspace*{.5cm} in($\langle$i, Buyer, Seller, 
contractProposal(Merchandise, Req\_amount, Price), T$\rangle$, \\
\hspace*{1.5cm}msgbox:getMessage(Seller)), \\ 
\hspace*{.5cm} in(Min, oracle:select(stored\_merchandise.max, name, =, Merchandise)),\\
\hspace*{.5cm} Price $<=$ Min
}\\
If the price proposed by the buyer is below the {\em Min} threshold, then the seller 
agent refuses the proposal. 

\item[] ~~\\
\noindent {\em 
{\bf Do} 
sendMessage(Seller, Buyer, refuse(Merchandise, Req\_amount, Price)) $\leftarrow$ \\
\hspace*{.5cm} in($\langle$i, Buyer, Seller, 
contractProposal(Merchandise, Req\_amount, Price), T$\rangle$, \\
\hspace*{1.5cm} msgbox:getMessage(Seller)), \\ 
\hspace*{.5cm} in(Amount, oracle:select(stored\_merchandise.amount, name, =, Merchandise)), \\
\hspace*{.5cm}  Amount $<$ Req\_amount
}\\
The proposal is refused if there is not enough merchandise available. 

\item[] ~~\\
\noindent {\em 
{\bf Do} 
sendMessage(Seller, Buyer, \\
\hspace*{.5cm} contractProposal(Merchandise, Req\_amount, Means)) $\leftarrow$ \\
\hspace*{.5cm} in($\langle$i, Buyer, Seller, 
contractProposal(Merchandise, Req\_amount, Price), T$\rangle$,
\\ 
\hspace*{1.5cm} msgbox:getMessage(Seller)), \\ 
\hspace*{.5cm} in(Max, oracle:select(stored\_merchandise.max, name, =, Merchandise)), \\
\hspace*{.5cm} in(Min, oracle:select(stored\_merchandise.min, name, =, Merchandise)),\\
\hspace*{.5cm} in(Amount, oracle:select(stored\_merchandise.amount, name, =, Merchandise)), \\
\hspace*{.5cm} Price $>$ Min, Price $<$ Max, Amount $>=$ Req\_amount\\
\hspace*{.5cm} in(Means, math:evalMeans(Max, Price))
}
\\
This rule manages the case the seller agent has to send a {\em contractProposal} 
back to the buyer, since the proposed price is between {\em Min} and {\em Max} and
there is enough merchandise available.
\end{itemize}


\section{Dynamic Logic}
\label{dynamic-logic}

Our introduction to dynamic logic is based on Section~8.2.5 of \cite{weiss99}.
Dynamic logic can be thought of as the modal logic of action. Unlike
traditional modal logics, the necessity and possibility operators of dynamic
logic are based upon the kinds of actions available. As a consequence of this
flexibility, dynamic logic has found use in a number of areas of Distributed Artificial
Intelligence (DAI). We consider the
propositional dynamic logic of regular programs, which is the most common
variant. This logic has a sublanguage based on regular expressions for defining
action expressions -- these composite actions correspond to Algol-60 programs,
hence the name of {\em regular programs}. 

Regular programs and formulae of the dynamic logic language
are defined by mutual induction.  
Let $PA$ be a set of atomic action symbols. 
Regular programs 
are defined in the following way: 
\begin{itemize}
\item all atomic action symbols in $PA$ are regular programs;
\item if $p$ and $q$ are regular programs, then $p~;q$ is a regular program
meaning {\em doing $p$ and $q$ in sequence};
\item if $p$ and $q$ are regular programs, then $(p+q)$ is a regular program 
meaning {\em doing either $p$ or $q$, whichever works};
\item if $p$ is a regular program, then $p*$ is a regular program 
meaning {\em repeating zero or more (but finitely many) iterations
of $p$};
\item if $\varphi$ is a formula of the dynamic logic language, 
then $\varphi?$ is a regular program representing
{\em the action of checking the truth value of formula} $\varphi$; 
it succeeds if $\varphi$ is indeed found to be true.
\end{itemize}
$(p+q)$ is nondeterministic choice. This action might sound a little
unintuitive since a nondeterministic program may not
be physically executable, requiring arbitrary lookahead to infer
which branch is really taken. From a logical viewpoint, however, the
characterization of nondeterministic choice is clear.
As far as $\varphi?$ is concerned, if $\varphi$ is true, this action succeeds as a {\em
noop}, i.e. without affecting the state of the world. If $\varphi$ is false, it fails,
and the branch of the action of which it is part terminates in failure --
it is as if the branch did not exist.
 
Dynamic logic formulae are defined in the following way:
\begin{itemize}
\item all propositional formulae are dynamic logic formulae;
\item if $p$ is a regular program and $\varphi$ is a 
dynamic logic formula, then $[p]\varphi$ is a dynamic logic formula
which means that 
{\em whenever $p$ terminates, it must do so in a state satisfying $\varphi$};
\item if $p$ is a regular program and $\varphi$ is a 
dynamic logic formula, then $\langle p \rangle \varphi$ is a dynamic logic formula
 which means that 
{\em it is possible to execute $p$ and halt in a state satisfying $\varphi$};
\item if $\varphi$ and $\psi$ are dynamic logic formulae, then 
$\varphi \vee \psi$, $\varphi \wedge \psi$, $\varphi \implies \psi$ and $\neg \varphi$ are
dynamic logic formulae.
\end{itemize}

Let  $S$ be a set of states (or worlds).
Let $h$ be an  interpretation function
\[h: S \times PROP \rightarrow \{\textrm{true}, \textrm{false}\} \]
which says if a propositional formula belonging to $PROP$
is true or false in a world belonging to $S$.  
Let $\sigma \subseteq S \times PA \times S$ be a transition relation. 

The semantics of dynamic logic is given with respect to a model 
$\langle S, \sigma, h \rangle$ that includes a
set $S$ of states, a transition relation $\sigma$ 
and an interpretation function $h$.

In order to provide the semantics of the language we first define a class of
accessibility relations ($\beta$ is a atomic action symbol from $PA$; 
$p$ and $q$ are regular programs;
$r$, $s$ and $t$, with subscripts when necessary, are members of $S$):
 
\begin{tabbing}
Spaziolo \= Formulalalalalalala \= *iff* \=  \kill 
\> $s ~R_{\beta}~ t $ \>
iff \> $ \sigma(s, \beta, t) $
\\
\> $ s ~R_{p;q}~ t$ \>
iff \> there exists $ r$ such that $s~ R_p ~r$ and $r ~R_q ~t$
\\
\> $ s ~R_{p+q}~ t$ \>
iff \> $s ~R_p~ t$ or $s~ R_q~ t$
\\
\> $ s~ R_{p*}~ t$ \>
iff \> there exists $ s_0, \ldots, s_n $ such that 
$s = s_0$ and $t = s_n$ \\
\> \> \> and for all $i, 0 \leq i < n, s_i ~R_p~ s_{i+1} $ 
\\
\> $ s~ R_{\varphi?}~ s$ \>
iff \> $ \langle S, \sigma, h \rangle \models_s \varphi  $ 
\end{tabbing}

In the following equivalences $p$ ranges over regular programs and
$\varphi$ and $\psi$ are dynamic logic formulae. 

If $\varphi$ is a propositional formula, its semantics is given 
through the $h$ interpretation function: 
\begin{tabbing}
Spaziolo \= Formulalalalalalalalalo \= *iff* \=  \kill 

\> $ \langle S, \sigma, h \rangle \models_s \varphi $ \>
iff \>  $h(s, \varphi) = true $ 
\end{tabbing}

The semantics of $\varphi \vee \psi$, $\varphi \wedge \psi$, 
$\varphi \implies \psi$ and $\neg \varphi$ is given in the standard way:
\begin{tabbing}
Spaziolo \= Formulalalalalalalalalo \= *iff* \=  \kill 
\> $ \langle S, \sigma, h \rangle \models_s \varphi \vee \psi $ \>
iff \> $ \langle S, \sigma, h \rangle \models_s \varphi
~~\textrm{or}~~\langle S, \sigma, h \rangle \models_s \psi $
\\
\> $ \langle S, \sigma, h \rangle \models_s \varphi \Rightarrow \psi $ \>
iff \> $ \langle S, \sigma, h \rangle \models_s \varphi
~~\textrm{implies}~~\langle S, \sigma, h \rangle \models_s \psi $ 
\\
\> etc...
\end{tabbing}

The semantics of $\langle p \rangle \varphi$ and $[ p ] \varphi$ is given as:
\begin{tabbing}
Spaziolo \= Formulalalalalalalalalo \= *iff* \=  \kill 
\> $ \langle S, \sigma, h \rangle \models_s \langle p \rangle \varphi $ \>
iff \>  there exists  $t$ such that $s~ R_p ~t$ and $ \langle S, \sigma, h
\rangle \models_{t} \varphi $ 
\\
\> $ \langle S, \sigma, h \rangle \models_s [ p ] \varphi $ \>
iff \>  for all  $t$, $s~ R_p ~t$ implies $ \langle S, \sigma, h
\rangle \models_{t} \varphi $ 
\end{tabbing}

The reader can refer to the survey \cite{kozen-tiurzyn} for additional details.

\subsection{\dylog}
\label{agent-logic-action-language}

In a set of papers \cite{martelli8,martelli9,martelli1,martelli2}, 
Baldoni, Giordano, Martelli, Patti and Schwind 
describe an action language and its extension to deal with complex actions. 
In this section we provide a short description of the action language, taken 
from \cite{BGMP}, and we introduce an implementation.
The implementation language is called  \dylog. 

\paragraph*{Primitive actions.}

In the action language each primitive action $a \in A$ is represented by a 
modality $[a]$. The meaning of the formula $[a]\alpha$, where $\alpha$ 
is an epistemic fluent, is that 
$\alpha$ holds after any execution of $a$. 
The meaning of the formula $\langle a \rangle \alpha$ 
is that there is a possible execution of action $a$ after which $\alpha$ holds. 
There is also a modality \nec\ which is used to denote those formulae holding 
in all states. A state consists in a set of {\em fluents} representing 
the agent's knowledge in that state. They are called fluents because their value 
may change from state to state. 
The {\em simple action laws} are rules that allow to describe direct 
{\em action laws}, {\em precondition laws} and {\em causal laws}. 

{\em Action laws} define the direct effect of primitive actions on a fluent and allow 
actions with conditional effects to be represented. They have the form 

\centerline{$\necm  (F_s \rightarrow [a]F)$}

where $a$ is a primitive action name, $F$ is a fluent, and $F_s$ is a fluent 
conjunction, meaning that action $a$ initiates $F$, when executed in a state 
where the {\em fluent precondition} $F_s$ holds.

{\em Precondition laws} allow {\em action preconditions}, i.e. those conditions 
which make an action executable in a state, to be specified. Precondition laws 
have form

 \centerline{$\necm  (F_s \rightarrow \langle a \rangle true)$}

meaning that when a fluent conjunction $F_s$ holds in a state, execution of the 
action $a$ is possible in that state.

{\em Causal laws} are used to express causal dependencies among fluents and, then, 
to describe {\em indirect} effects of primitive actions. They have the 
form 

\centerline{$\necm  (F_s \rightarrow F)$}
meaning that the fluent $F$ holds if the fluent conjunction $F_s$ holds too.

In the implementation language \dylog\ the notation for the above constructs is
the following: action 
laws have the form 
\emph{a causes F if Fs}, precondition laws have the form 
\emph{a possible\_if Fs}
and causal laws have the form \emph{F if Fs}.

\paragraph*{Procedures.}

{\em Procedures}  are 
defined on the basis of primitive actions, test 
actions and other procedures. Test actions are needed 
for testing if some fluent holds in the current 
state and for expressing conditional procedures and are written as 

\centerline{$F_s ?$}

where $F_s$ is a fluent conjunction. 

A procedure $p_0$ is defined by means of a set of 
inclusion axiom schemas of the form 

\centerline{$\langle p_1 \rangle \langle p_2 \rangle \ldots \langle p_n \rangle 
\varphi
\Rightarrow \langle p_0 \rangle \varphi$}

where $\varphi$ stands for an arbitrary formula. 

In \dylog\ implementations a {\em procedure} is defined as a collection 
of {\em procedure clauses} 
of the form
\emph{$p_0$ isp $p_1$ \& \ldots \& $p_n$} $(n \geq 0)$
where \emph{$p_0$} is the name of the procedure and \emph{$p_i$}, 
$i = 1 \ldots n$ is either a
 primitive action, a test action (written $F_s ?$), a procedure name, or a \prolog\ goal. 
 Procedures can be recursive and they are executed in a goal directed way, 
 similarly to standard logic programs. 

\paragraph*{Planning.}

A {\em planning problem} amounts to determining, given an initial state and a goal
 $F_s$, if there is a possible execution of a procedure $p$
 leading to a state in which $F_s$ holds. This can be formulated by the query 

\centerline{$\langle p \rangle F_s$} 

The execution of the above query 
returns as a side effect an answer which is an {\em execution trace} 
$ a_1, a_2, \ldots, a_m $, i.e. a primitive action sequence from the initial state 
to the final one, which represents a {\em linear plan}. 

To achieve this, \dylog\ provides a metapredicate 
\emph{plan(p, Fs, as)}
where \emph{p} is a procedure, \emph{Fs} a goal and \emph{as} a sequence of 
primitive actions. 
\emph{plan} simulates the execution of the procedure \emph{p}.
If \emph{p} includes sensing actions, the possible execution traces of  \emph{p}
(represented by sequences \emph{as} of primitive actions) are generated
according to the possible outcomes of the sensing actions. When one 
(simulated) outcome leads
to a final state where \emph{Fs} is not satisfied, the interpreter backtracks
and alternative outcome is simulated. 
Execution is separated from planning: a metapredicate 
{\em exe(as)} is provided to execute a plan.

\paragraph*{Sensing.} In general, it is not possible to assume that the value of 
each fluent in a state is known to an agent, and it is necessary to represent the 
fact that some fluents are unknown and to reason about the execution of actions on 
incomplete states. To represent explicitly the unknown value of some fluents, 
an {\em epistemic operator} ${\bf B}$ is introduced in the language, to represent 
the beliefs an agent has on the world. ${\bf B} f$ will mean that the fluent $f$ 
is known to be true, ${\bf B} \neg  f$ will mean that the fluent $f$ is known to 
be false, and fluent $f$ is undefined in the case both $\neg {\bf B} f$ and 
$\neg {\bf B} \neg  f$ hold. In the following, 
$u(f)$ stands for $\neg {\bf B} f ~ \wedge ~ \neg {\bf B} \neg  f$. 

In \dylog\ there is no explicit use of the operator ${\bf B}$ but the notation is 
extended with the test \emph{u(f)?}. Thus each fluent can have one of the three 
values: true, false and unknown. An agent will be able to know the value of 
\emph{f} 
by executing an action that senses \emph{f} ({\em sensing} action).
One expresses that an action \emph{s} causes to know whether \emph{f} holds
by the declaration \emph{s senses f}.
By applying \dylog's planning predicate 
\emph{plan} to a procedure containing sensing actions a {\em conditional plan} 
is obtained. The branches of this plan correspond to the different outcomes of 
sensing actions.

\subsubsection{Semantics}

As discussed in \cite{martelli2},  the  logical characterization of \dylog\
can be provided in two steps. First, a multimodal logic 
interpretation of a dynamic domain description which describes the monotonic part 
of the language is introduced. Then, an abductive semantics to 
account for non-monotonic behavior of the language is provided.

\begin{definition}[Dynamic domain description]
{\rm
Given a set $\calA$ of atomic world actions, a set $\calS$ of sensing actions, and a
set $\calP$ of procedure names, let $\Pi_\calA$ be a set of simple action laws for
world actions, $\Pi_\calS$ a set of axioms for sensing actions, and
$\Pi_\calP$ a set of inclusion axioms. A {\em dynamic domain 
description} is a pair $(\Pi, S_0)$, where $\Pi$ is the tuple $(\Pi_\calA, \Pi_\calS, 
\Pi_\calP)$ and $S_0$ is a consistent and complete set of epistemic literals representing 
the beliefs of the agent in the initial state.
}
\end{definition}

\paragraph*{Monotonic interpretation of a dynamic domain description.}

Given a dynamic domain description $(\Pi, S_0)$, let us call $\calL_{(\Pi, S_0)}$ the
propositional modal logic on which $(\Pi, S_0)$ is based. The action laws for primitive
actions in $\Pi_{\calA}$ and the initial beliefs in $S_0$ define a theory   
$\Sigma_{(\Pi, S_0)}$ in $\calL_{(\Pi, S_0)}$.
The axiomatization of $\calL_{(\Pi, S_0)}$, called $\calS_{(\Pi, S_0)}$, contains:
\begin{itemize}
\item all the axioms for normal modal operators
\item $D({\bf B})$, namely, for the belief modality 
${\bf B}$ the axiom ${\bf B} p \implies \neg {\bf B} \neg p$ 
holds; 
\item $ S4(\Box)$, namely, the three axioms $\Box p \implies p$, 
$\Box (p \implies q) \implies (\Box p \implies \Box q)$ and \\
$\Box p \implies \Box \Box p$;
\item $ \Box \varphi \implies [a_i]\varphi$, one for each primitive action $a_i$ in
$(\Pi, S_0)$;
\item $\langle a + b \rangle \varphi \equiv  \langle a \rangle \varphi \vee  \langle b \rangle \varphi$, one for each formula $\varphi$;
\item $\langle \psi? \rangle \varphi \equiv \psi \wedge \varphi$, one for each formula $\varphi$;
\item $\langle a;b \rangle \varphi \equiv \langle a \rangle \langle b \rangle \varphi$, one for each formula $\varphi$;
\item $\Pi_{\calP}$;
\item $\Pi_{\calS}$.
\end{itemize}
The model theoretic semantics of the logic $\calL_{(\Pi, S_0)}$ is given through a 
standard Kripke semantics with inclusion properties among the accessibility relations.
More details can be found in \cite{baldoni98}.

\paragraph*{Abductive semantics.}

The monotonic part of the language does not account for persistency. 
In order to deal with the frame problem, it is necessary to
 introduce a non-monotonic semantics 
for the language by making use of an abductive construction: abductive assumptions 
will be used to model persistency from one state to the following one, when a 
primitive action is performed. In particular, we will assume that a fluent expression
 $F$ persists through an action unless it is inconsistent to assume so, i.e. unless
 $\neg F$ holds after the action.

In defining the abductive semantics, the authors adopt (in a modal setting) 
the style of Eshghi and Kowalski's abductive 
semantics for negation as failure \cite{eshghikowalski89}. 
They introduce the notation ${\bf M}\alpha$ to denote a new atomic 
proposition associated with $\alpha$
and a set of atomic propositions 
of the form 
${\bf M}[a_1][a_2]...[a_m]F$ and take them as 
being abducibles\footnote{${\bf M}$ is not a modality but just a
notation adopted in analogy to default logic, where a justification 
${\bf M}\alpha$ intuitively means ``$\alpha$ is consistent''.}. Their meaning 
is that the fluent expression $F$ can be assumed to hold in the state obtained 
by executing primitive actions $a_1, a_2, ..., a_m$. Each abducible can be assumed to 
hold, provided it 
is consistent with the domain description $(\Pi, S_0)$ and with other assumed abducibles. 

More precisely, in order to deal with the frame problem, they add to the axiom 
system of $\calL(\Pi, S_0)$ the {\em persistency axiom schema}

\centerline{$[a_1][a_2]...[a_{m-1}]F \wedge {\bf M}[a_1][a_2]...[a_{m-1}][a_m]F 
\implies [a_1][a_2]...[a_{m-1}][a_m]F$}
 where $a_1, a_2, ...., a_m (m > 0)$ are primitive actions, and 
$F$ is a fluent expression. Its meaning is that, if $F$ holds after action sequence 
$a_1, a_2,..., a_{m-1}$, and $F$ can be assumed to persist after action 
$a_m$ (i.e., it is consistent to assume ${\bf M}[a_1][a_2]...[a_m]F$), then we can 
conclude that $F$ holds after performing the sequence of actions 
$a_1, a_2,..., a_m$.

Besides the  persistency action schema, the authors provide the notions of abductive
solutions for a dynamic domain description and abductive solutions to a 
query. 

\subsubsection{Implementation}

\dylog\ is defined by a 
proof procedure which constructs a linear plan by making assumptions on the possible 
results of sensing actions. The goal directed proof procedure, based on negation as
failure, allows a query to be proved from a given dynamic domain description.
The proof procedure is sound and complete with respect to the Kripke semantics
of the modal logics $\calL(\Pi, S_0)$. An interpreter based on this proof procedure
has been implemented in \sicstus. This implementation allows to use
\dylog\ as a programming language for executing procedures which model the behavior of
an agent, but also to reason about them, by extracting from them linear or 
conditional plans. Details on the implementation can be found in \cite{dylogDownload}.

\subsubsection{Extensions}

In \cite{reasoningAboutConvProt,reasoningAboutSelfAndOthers} \dylog\ agents are 
extended to represent beliefs of
other agents in order to reason about conversations.
They are also enriched with a communication kit
including a primitive set of speech acts, a set of special ``get message'' actions and
a set of conversation protocols.

\subsubsection{Example}
In the following example, the predicate {\em is} has its usual meaning as in \prolog\ 
programs: it evaluates the value of the expression at its right
and checks if this value unifies with the term at its left. 

\begin{itemize}
\item Functional fluents:
	\begin{itemize}
	\item[]  {\em functionalFluent(storing/2)}.
	\item[]  {\em functionalFluent(new\_message/2)}.
	\item[] The amount of merchandise stored and the new incoming messages 
	are facts which change during the agent's life. The number after the 
	predicate's name is its arity.
	\end{itemize}
\item Unchangeable knowledge base (\prolog\ facts):
	\begin{itemize}
	\item[] {\em min-price(orange, 1)}.
	\item[] {\em max-price(orange, 2)}.
	\item[] The minimum and maximum prices for oranges do not change over time.
	\end{itemize}
\item Initial observations:
	\begin{itemize}
	\item[] {\em obs(storing(orange, 1000))}.
	\item[] Initially, there are 1000 oranges the seller agent can sell.
	\end{itemize}

\item Primitive actions:
       \begin{itemize}
	\item[] {\em receive}\\
This action senses if a fluent {\em new\_message(Sender,  Message)} is
present in the caller's mailbox.
It is characterized
by the following laws and routines:
		\begin{itemize}
		\item[] Precondition laws:\\
{\em receive possible\_if true.}\\
It is always possible to wait for a new message to arrive.
	\item[] Sensing:\\
 {\em receive  senses new\_message(Sender, Message).}\\
The {\em receive} action senses the value of the {\em 
new\_message(Sender, Message)} functional fluent.
\item[] Sensing routine:\\
{\em senses\_routine(\_, new\_message, Sender, Message) :- ....}\\
A \prolog\ routine that we do not show here 
 implements the sensing action by waiting 
for messages matching the couple {\em (Sender, Message)} in the 
caller's mailbox.
\end{itemize}
	\item[] {\em send(Sender, Receiver, Message)}\\
This action puts the couple {\em (Sender,  Message)} in the {\em Receiver}'s mailbox
by modifying the state of the {\em 
new\_message(Sender, Message)} functional fluent of the {\em Receiver}'s agent.
For sake of conciseness, we avoid discussing all the details of this action.
\item[] {\em ship(Buyer, Merchandise, Req\_Amnt, Price)}\\
This action ships the required merchandise to the {\em Buyer} agent. It is characterized
by the following action laws and precondition laws:
		\begin{itemize}
		\item[] Action laws:\\
		{\em ship(Buyer, Merchandise, Req\_Amnt, Price) \\
\hspace*{.5cm} causes storing(Merchandise, Amount) \\
\hspace*{.5cm} if storing(Merchandise, Old\_Amount) \& \\ 
\hspace*{1cm} (Amount is Old\_Amount - Req\_Amnt)}.\\
Shipping some merchandise causes an update of the stored amount of that merchandise.
		\item[] Precondition laws:\\
{\em ship(Buyer, Merchandise, Req\_Amnt) \\
\hspace*{.5cm} possible\_if  storing(Merchandise, Old\_Amount) \& \\ 
\hspace*{1cm} (Old\_Amount $>=$ Req\_Amnt) \& \\
\hspace*{1cm} max-price(Merchandise, Max) \&
 (Price $>=$ Max)}.\\
Shipping merchandise is possible if there is enough merchandise left and
if the price is higher than the maximum price 
established for that merchandise.
		\end{itemize}
	\end{itemize} 

\item Procedures:
	\begin{itemize}
	\item[] {\em seller\_agent\_cycle isp\\
	\hspace*{.5cm} receive  \&\\
	\hspace*{.5cm} manage\_message \&\\
	\hspace*{.5cm} seller\_agent\_cycle}.\\
The main cycle for the seller agent consists in waiting for  a message,
 managing it and starting waiting for a message again.
\item[] \vspace*{.3cm} {\em manage\_message isp \\
	\hspace*{.5cm} new\_message(Buyer, \\
\hspace*{1cm} contractProposal(Merchandise, Req\_Amnt, Price))? \& \\
\hspace*{.5cm} storing(Merchandise, Old\_Amount)? \& \\ 
\hspace*{.5cm} (Old\_Amount $>=$ Req\_Amnt) \& \\
\hspace*{.5cm} max-price(Merchandise, Max)? \& (Price $>=$ Max) \&\\
\hspace*{.5cm} ship(Buyer, Merchandise, Req\_Amnt, Price) \& \\
\hspace*{.5cm} send(seller, Buyer,
accept(Merchandise, Req\_Amnt, Price))
}\\
If all conditions are met to ship the merchandise, then the merchandise is shipped and the
seller sends a message to the {\em Buyer} in which it accepts the {\em Buyer}'s proposal. 

\item[] \vspace*{.3cm} {\em manage\_message isp \\
	\hspace*{.5cm} new\_message(Buyer, \\
\hspace*{1cm} contractProposal(Merchandise, Req\_Amnt, Price))? \& \\
\hspace*{.5cm} storing(Merchandise, Old\_Amount)? \& \\ 
\hspace*{.5cm} (Old\_Amount $<$ Req\_Amnt) \& \\
\hspace*{.5cm} send(seller, Buyer, 
refuse(Merchandise, Req\_Amnt, Price))
}\\
If there is not enough merchandise, the seller agent refuses to send the merchandise.

\item[] \vspace*{.3cm} {\em manage\_message isp \\
	\hspace*{.5cm} new\_message(Buyer, \\
\hspace*{1cm} contractProposal(Merchandise, Req\_Amnt, Price))? \& \\
\hspace*{.5cm} min-price(Merchandise, Min)? \& (Price $<=$ Min) \\
\hspace*{.5cm} send(seller, Buyer, 
refuse(Merchandise, Req\_Amnt, Price))
}\\
If the price is too low, the seller agent refuses to send the merchandise.

\item[] \vspace*{.3cm} {\em manage\_message isp \\
	\hspace*{.5cm} new\_message(Buyer, \\
\hspace*{1cm} contractProposal(Merchandise, Req\_Amnt, Price))? \& \\
\hspace*{.5cm} storing(Merchandise, Old\_Amount)? \& \\ 
\hspace*{.5cm} (Old\_Amount $>=$ Req\_Amnt) \&  \\
\hspace*{.5cm} max-price(Merchandise, Max)? \& (Price $<$ Max) \&\\
\hspace*{.5cm} min-price(Merchandise, Min)? \& (Price $>$ Min) \&\\
\hspace*{.5cm} (New\_Price is (Price + Max) / 2) \& \\
\hspace*{.5cm} send(seller, Buyer, 
contractProposal(Merchandise, Req\_Amnt, New\_Price))}\\
In case  the conditions are met to send a counter-proposal to the {\em Buyer} agent,
the seller sends it with the price it is willing to accept. 

	\end{itemize}
\end{itemize}
\section{Temporal Logic}
\label{temporal-logic}

In this section we define a first-order temporal logic
based on discrete, linear models with finite past and infinite future, called
{\em FML} \cite{fisherNormalForm}. 
FML introduces two new connectives to classical logic, {\em until} (\until) and
{\em since} (\since), together with a number of other operators definable 
in terms of \until\ and  \since. The intuitive meaning
of a temporal logic formula 
$\varphi$ \until $\psi$ is that $\psi$ will
become  true at some future time point $t$ and  that in all states
between and different from now and $t$, $\varphi$ will be true. 
\since\ is the analogous of \until\ in the past. 

\paragraph*{Syntax of FML.}

{\em Well-formed formulae} of FML (WFF$_f$) are generated in the usual way as for
classical logic, starting from a set ${\cal L}_p$ of predicate symbols, a set ${\cal L}_v$
of variable symbols, a set ${\cal L}_c$ of constant symbols, the quantifiers $\forall$ and
$\exists$, and the set ${\cal L}_t$ of terms (constants and variables).
The set WFF$_f$ is defined by:
\begin{itemize}
\item If $t_1, ..., t_n$ are in ${\cal L}_t$ and $p$ is a predicate symbol of arity $n$, 
then $p(t_1, ..., t_n)$ is in WFF$_f$.
\item {\em true} and {\em false}  are in WFF$_f$. 
\item If $A$ and $B$ are in WFF$_f$, then so are  
$\neg A$, $A \wedge B$, $A$ \until $B$, $A$ \since $B$, and $(A)$.
\item If $A$ is in WFF$_f$  and $v$ is in ${\cal L}_v$, then $\exists v. A$ and $\forall v. A$ 
are both in WFF$_f$.
\end{itemize}  

The other classical connectives are defined in terms of the ones given above, 
and several other useful temporal connectives are defined in terms of ${\cal U}$ and
${\cal S}$ (we follow the characterization provided in \cite{finger-fisher-owens}
as well as the notation used there):
\begin{tabbing}
Spazio \= Formula \= ExplanationOfTheMeaningCanBeVeryLongLongLong \=  \kill 
\> \nextt $\varphi$ \> {\small $\varphi$ is true in the next state} \> [{\em false} \until $\varphi$] \\
\> \stronglast $\varphi$ \> {\small the current state is not the initial state, and $\varphi$ was true in} \\[-.15cm]
\> \> {\small the previous
state} \> [{\em false} \since $\varphi$] \\
\> $\weaklast \varphi$ \> {\small if the current state is not the initial state, then $\varphi$ was true in} \\[-.15cm]
\> \> {\small the previous
state} \> [$\neg$ \stronglast $\neg \varphi$] \\
\> \poss $\varphi$ \> {\small $\varphi$ will be true in some future state} \>
 [{\em true} \until $\varphi$] \\
 \> $\was \varphi$ \> {\small $\varphi$ was true in some past state} \> 
 [{\em true} \since $\varphi$] \\
 \> \nec $\varphi$ \> {\small $\varphi$ will be true in all future states} 
 \> [$\neg$ \poss $\neg \varphi$] \\
 \> $\heretofore \varphi$ \> {\small $\varphi$ was true in all past states} \>  
 [$\neg \was \neg \varphi$] \\
\end{tabbing}
Temporal formulae can be classified as follows. A {\em state-formula} is either a literal or a 
boolean combination of other state-formulae. 
\\
{\em Strict future-time}
formulae are defined as follows:
\begin{itemize}
\item[] If $A$ and $B$ are either state or strict future-time formulae, 
then $A$ \until $B$ is a strict future-time formula.
\item[] If $A$ and $B$ are strict future-time formulae,
then $\neg A$, $A \wedge B$, and $(A)$ are strict future-time formulae. 
\end{itemize}
{\em Strict past-time} formulae are defined as the past-time duals of 
strict future-time formulae. 
\\
{\em Non-strict} classes of formulae include state-formulae in their  definition.

\paragraph*{Semantics of FML.}

The models for FML formulae are given by a {\em structure} which consists of a sequence of 
states, together with an {\em assignment} of truth values to atomic sentences
within states, a domain ${\cal D}$ which is assumed to be constant for every
state, and mappings from elements of the language into denotations.
More formally, a model is a tuple ${\cal M} = \langle \sigma, {\cal D}, h_c, h_p \rangle$ where
$\sigma$ is the ordered set of states $s_0, s_1, s_2, ...$, $h_c$ is a map from
the constants into ${\cal D}$, and $h_p$ is a map from ${\bf N} \times {\cal L}_p$ 
into ${\cal D}^n \rightarrow$ \{ true, false \} (the first argument of 
$h_p$ is the index $i$ of the state $s_i$). 
Thus, for a particular state $s$, and a
particular predicate $p$ of arity $n$, $h(s, p)$ gives truth values to atoms 
constructed from $n$-tuples of elements of ${\cal D}$. 
A {\em variable} assignment $h_v$ is a mapping from the variables into 
elements of ${\cal D}$. Given a variable and the valuation function $h_c$, a term
assignment $\tau_{vh}$ is a mapping from terms into ${\cal D}$ defined in the usual way.

The semantics of FML is given by the $\models$ relation that gives the truth value
of a formula in a model ${\cal M}$ at a particular moment in time $i$ and
with respect to a variable assignment.

\begin{tabbing}
Sp \= Formulalalalalalalalalolololo \= *iff* \=  \kill 
\> $\langle {\cal M}, i, h_v \rangle \models$ {\em true} \> \> 
\\
\> $\langle {\cal M}, i, h_v \rangle \not\models$ {\em false} \> \> 
\\
\> $\langle {\cal M}, i, h_v \rangle \models p(x_1, ..., x_n) $ \> iff 
\> $h_p(i, p)(\tau_{vh}(x_1),..., \tau_{vh}(x_n)) =$ {\em true}  
\\
\> $\langle {\cal M}, i, h_v \rangle \models \neg \varphi $ \> iff 
\> $\langle {\cal M}, i, h_v \rangle \not\models \varphi $ 
\\
\> $\langle {\cal M}, i, h_v \rangle \models \varphi \vee \psi $ \> iff 
\> $\langle {\cal M}, i, h_v \rangle \models \varphi~~\textrm{or}~~ 
\langle {\cal M}, i, h_v \rangle \models \psi$ 
\\
\> $\langle {\cal M}, i, h_v \rangle \models \varphi$ \until $\psi $ \> iff 
\> for some $k$ such that $i < k$, $\langle {\cal M}, k, h_v \rangle \models \psi$\\
\> \> \> 
and for all $j$, if $i < j < k$ then
$\langle {\cal M}, j, h_v \rangle \models \varphi$ 
\\
\> $\langle {\cal M}, i, h_v \rangle \models \varphi$ \since $\psi $ \> iff 
\> for some $k$ such that $0 \leq k < i$, $\langle {\cal M}, k, h_v \rangle \models \psi$\\
\> \> \>
and for all $j$, if $k < j < i$ then
$\langle {\cal M}, j, h_v \rangle \models \varphi$ 
\\
\> $\langle {\cal M}, i, h_v \rangle \models \forall x. \varphi $ \> iff 
\> for all $d \in {\cal D}, \langle {\cal M}, i, h_v[d/x] \rangle \models \varphi$ 
\\
\> $\langle {\cal M}, i, h_v \rangle \models \exists x. \varphi $ \> iff 
\> there exists $d \in {\cal D}$ such that $\langle {\cal M}, i, h_v[d/x] \rangle \models \varphi$ 
\\
\end{tabbing}

\subsection{\concMTM}
\label{concMTM}

\concMTM\ \cite{concMTM14,concMTM13,concMTM16} is a programming language for 
distributed artificial intelligence based on FML.
A \concMTM\ system contains a number of concurrently executing agents which are
able to 
communicate through message passing. Each agent executes a first-order temporal logic
specification 
of its desired behavior. 
Each agent has two main components:
\begin{itemize}
\item an {\em interface} which defines how the agent may interact with its
environment 
(i.e., other agents);
\item a {\em computational engine}, which defines how the agent may act.
\end{itemize}

An agent interface consists of three components:

\begin{itemize}
\item a unique {\em agent identifier} which names the agent
\item a set of predicates defining what messages will be accepted by the agent -- 
they are called {\em environment predicates};
\item a set of predicates defining messages that the agent may 
send -- these are called {\em component predicates}. 
\end{itemize}
Besides environment and component predicates, an agent has a set of 
{\em internal predicates} with no external effect.

The computational engine of an object is based on the \MTM\ paradigm of
executable 
temporal logics. The idea behind this approach is to directly execute a
declarative 
agent specification given as a set of {\em program rules} which are temporal
logic 
formulae of the form:

\centerline{antecedent about past $\Rightarrow$ consequent about future} 

The past-time antecedent is a temporal logic formula referring strictly to the
past, 
whereas the future time consequent is a temporal logic formula referring either
to 
the present or future. The intuitive interpretation of such a rule is {\em on
the 
basis of the past, do the future}. The individual \MTM\ rules are given in 
the FML logic defined before. Since \MTM\ rules must respect the {\em past implies
future} form, FML formulae defining agent rules must be transformed into this
form. This is always possible as demonstrated in \cite{BFG89}.  

\subsubsection{Semantics}

\MTM\ semantics is the one defined for FML.  

\subsubsection{Implementation}

Two implementations of the imperative future paradigm described in this section
have been produced. The first is a prototype interpreter for propositional 
\MTM\ implemented in \textsf{Scheme} \cite{scheme}. 
A more robust \prolog-based interpreter for a restricted first-order version of
\MTM\ has been used as a transaction programming language for temporal databases 
\cite{finger91}.  

\subsubsection{Extensions}
\label{concExtensions}

Two main directions have been followed in the attempt
of extending \concMTM, the first one dealing with 
single agents and the second one dealing with MASs.
\begin{itemize}
\item Single \concMTM\ agents have been extended with deliberation and beliefs
\cite{michaelIJCAI97} and with resource-bounded reasoning \cite{michaelIJCAI99}.
\item Compilation techniques for MASs specified in \concMTM\ are analyzed in
\cite{michaelTIME97}. \concMTM\ has been proposed as a coordination language 
in \cite{michaelCOORDINATION97}. The definition of groups of agents in \concMTM\ 
is discussed in \cite{michaelATAL98,michaelISLIP00}.
\end{itemize}
The research on single \concMTM\ agents converged with the research on \concMTM\ MASs
in the paper \cite{michaelAAMAS02} where ``confidence'' is added to both single 
and multiple agents. The  development of teams of agents is 
discussed in \cite{michaelFAABS02}.

\subsubsection{Example}

The \concMTM\ program for the seller agent may be as follows:
\begin{itemize}
\item The interface of the {\em seller} agent is the following:	
	\begin{itemize}
	\item[] {\em seller(contractProposal)[accept, refuse, contractProposal, ship]}
	\\ meaning that:\\
        \noindent -- the seller agent, identified by the {\em seller} identifier, 
        is able to 
	recognize a {\em contractProposal} message with its arguments, not specified in 
        the interface; \\
        \noindent -- the messages that the seller agent is able
	to broadcast to the environment, including both communicative
	acts and actions on the environment, are {\em accept, refuse, 
        contractProposal, ship} with their arguments.
	\end{itemize}
\item The internal knowledge base of the seller agent contains the following
{\em rigid} predicates (predicates whose value never changes):
\begin{itemize}
	\item[] {\em min-price(orange, 1)}.
	\item[] {\em max-price(orange, 2)}. 
\end{itemize}
\item The internal knowledge base of the seller agent contains the following
{\em flexible} predicates (predicates whose value changes over time):
\begin{itemize}
	\item[] {\em storing(orange, 1000)}.
\end{itemize}
\item The program rules of the seller agent are the following ones
(as usual, lowercase symbols are constants and uppercase ones are variables):
	\begin{itemize}
	\item[] {\em $\forall$  Buyer, Merchandise, Req\_Amnt, Price. \\
\hspace*{.5cm} \stronglast [contractProposal(Buyer, seller, Merchandise, Req\_Amnt, Price) $\wedge$ \\
\hspace*{.5cm} storing(Merchandise, Old\_Amount) $\wedge$ \\ 
\hspace*{.5cm} Old\_Amount $>=$ Req\_Amnt $\wedge$ \\
\hspace*{.5cm} max-price(Merchandise, Max) $\wedge$ Price $>=$ Max] $\implies$\\
\hspace*{1cm} [ship(Buyer, Merchandise, Req\_Amnt, Price) $\wedge$ \\
\hspace*{1cm} accept(seller, Buyer, Merchandise, Req\_Amnt, Price)]
}\\
If there was a previous state where {\em Buyer} sent a {\em contractProposal}
message to {\em seller}, and in that previous state all the conditions 
were met to accept the proposal, 
then accept the {\em Buyer}'s proposal and ship the required merchandise. 
\item[] \vspace*{.3cm} {\em $\forall$  Buyer, Merchandise, Req\_Amnt, Price. \\
\hspace*{.5cm} \stronglast [contractProposal(Buyer, seller, Merchandise, Req\_Amnt, Price) 
$\wedge$ \\
\hspace*{.5cm} storing(Merchandise, Old\_Amount) $\wedge$ \\ 
\hspace*{.5cm} min-price(Merchandise, Min) $\wedge$ \\ 
\hspace*{.5cm} Old\_Amount $<$ Req\_Amnt $\vee$ Price $<=$ Min] $\implies$\\
\hspace*{1cm}  refuse(seller, Buyer, Merchandise, Req\_Amnt, Price)
}\\
If there was a previous state where {\em Buyer} sent a {\em contractProposal}
message to {\em seller}, and in that previous state the conditions were not 
met to accept the {\em Buyer}'s proposal, then send a {\em refuse} message
 to {\em Buyer}.
\item[] \vspace*{.3cm} {\em $\forall$  Buyer, Merchandise, Req\_Amnt, Price. \\
\hspace*{.5cm} \stronglast [contractProposal(Buyer, seller, Merchandise, Req\_Amnt, Price) 
$\wedge$ \\
\hspace*{.5cm} storing(Merchandise, Old\_Amount) $\wedge$ \\ 
\hspace*{.5cm} min-price(Merchandise, Min) $\wedge$ \\ 
\hspace*{.5cm} max-price(Merchandise, Max) $\wedge$ \\ 
\hspace*{.5cm} Old\_Amount $>=$ Req\_Amnt $\wedge$ \\
\hspace*{.5cm} Price $>$ Min $\wedge$ Price $<$ Max $\wedge$\\
\hspace*{.5cm} New\_Price $=$ (Max + Price) / 2] $\implies$\\
\hspace*{1cm}  contractProposal(seller, Buyer, Merchandise, Req\_Amnt, New\_Price)
}\\
If there was a previous state where {\em Buyer} sent a {\em contractProposal}
message to {\em seller}, and in that previous state the conditions were 
met to send a {\em contractProposal} back to {\em Buyer},
then send  a {\em contractProposal} message to {\em Buyer} with a new proposed price.
	\end{itemize}
\end{itemize}

\section{Linear Logic}
\label{linear-logic}

Linear logic \cite{Girard87} has been introduced as a resource-oriented 
refinement of classical logic. The idea behind linear logic is to constrain 
the number of times a given assumption (resource occurrence) can be used 
inside a deduction for a given goal formula. This resource 
management, together with the possibility of naturally modeling the notion 
of state, makes linear logic an appealing formalism to reason about concurrent 
and dynamically changing systems.

Linear logic extends usual logic with new connectives:
\begin{itemize}
\item {\em Exponentials}: ``!'' ({\em of course}) and ``?'' ({\em why not?})
 express the capability of an action of being iterated, i.e. the 
absence of any reaction. $!A$ means 
infinite amount of resource $A$.

\item {\em Linear implication}: $\lolli$ ({\em lolli}) is used for causal implication.
The relationship between linear implication and intuitionistic implication 
``$\Rightarrow$'' is $A \Rightarrow B \equiv (!A) \lolli B$  

\item {\em Conjunctions}: $\otimes$ ({\em times}) and $\&$ ({\em with})
 correspond to radically different uses of the word ``and''. Both conjunctions 
express the availability of two actions; but in the case of $\otimes$, both 
actions will be done, whereas in the case of $\&$ only one of them will be 
performed (we shall decide which one). Given an action of type $A \lolli B$ 
and an action of type $A \lolli C$
there will be no way of forming an action of type $A \lolli B \otimes C$, since 
once resource A has been consumed for deriving B, for example, it is not 
available for deriving $C$. However, there will be an action $A \lolli B \& C$. 
In order to perform this action we have to first choose which among the two 
possible actions we want to perform, and then do the one selected. 

\item {\em Disjunctions}: there are two disjunctions in linear logic, $\oplus$ 
({\em plus}), which is the dual of $\&$, and $\para~$ ({\em par}), which is the dual of 
$\otimes$.
$\oplus$ expresses the choice of one action between two possible types. $\para~$ 
expresses a dependency between two types of actions and can be used to model 
concurrency. 

\item {\em Linear negation}: $(\cdot)^{\perp}$ ({\em nil}) expresses linear negation. 
Since linear implication will eventually be rewritten as $A^{\perp} \para B$, 
{\em nil} is the only negative operation of logic. Linear 
negation expresses a {\em duality} 

\centerline{{\em action of type $A$ = reaction of type $A^{\perp}$}}

\item {\em Neutral elements}: there are four neutral elements: {\bf 1} 
(w.r.t. $\otimes$), $\perp$ (w.r.t. $\para$), $\top$ (w.r.t. $\&$) and {\bf 0} 
(w.r.t. $\oplus$).

\end{itemize}

\paragraph*{Semantics.} 

In Table \ref{system_for_LL} we provide the semantics of full linear
logic by means of a proof system. Sequents assume the one-sided, dyadic form 
$\dedllT{\Gamma}$. $\Theta$ and $\Gamma$ are multisets of formulae. 
$\Theta$ is the so-called {\em unbounded} part, while $\Gamma$ is the {\em bounded} one. 
In other words, 
formulae in $\Theta$ must be implicitly considered as exponentiated (i.e., preceded by $?$) 
and thus can be reused any number of times, while formulae in $\Gamma$ must be used exactly 
once. 

We opted for defining the semantics of linear logic
by means of a proof system both because 
understanding the proof rules requires less background than 
understanding an abstract semantics and for consistency with the 
style used for the semantics of \ehhf.
Other semantics have been defined for linear logic: 
a complete description of {\em phase semantics} and {\em coherent semantics} 
is given in \cite{Girard87} 
while {\em game semantics} is dealt with in \cite{blass}.

\begin{table*}
{\small $$
\begin{array}{ccc}
\infer[id]{\dedllT{F,\lneg{F}}}{} ~~&~~
\infer[cut]{\dedllT{\Gamma,\Delta}}{\dedllT{\Gamma,F} ~~&~~
                                   \dedllT{\Delta,\lneg{F}}} ~~&~~
\infer[abs]{\dedll{\Theta,F}{\Gamma}}{\dedll{\Theta,F}{\Gamma,F}}\\\\
\infer[\anti]{\dedllT{\Gamma,\anti}}{\dedllT{\Gamma}} ~~&~~
\infer[\para]{\dedllT{\Gamma,F\para G}}{\dedllT{\Gamma,F,G}} ~~&~~
\infer[?]{\dedllT{\Gamma,?F}}{\dedll{\Theta,F}{\Gamma}}\\\\
\infer[\one]{\dedllT{\one}}{} ~~&~~
\infer[\tensor]{\dedllT{\Gamma,\Delta,F\tensor G}}{\dedllT{\Gamma,F} ~~&~~
                                                  \dedllT{\Delta,G}} ~~&~~
\infer[!]{\dedllT{!F}}{\dedllT{F}}\\\\
\infer[\all]{\dedllT{\Gamma,\all}}{} ~~&~~
\infer[\with]{\dedllT{\Gamma,F\with G}}{\dedllT{\Gamma,F} ~~&~~
                                       \dedllT{\Gamma,G}} ~~&~~
\infer[\forall]{\dedllT{\Gamma,\forall x.F}}{\dedllT{\Gamma,\subst{F}{c}{x}}}
\\\\
\infer[\oplus_l]{\dedllT{\Gamma,F\oplus G}}{\dedllT{\Gamma,F}} ~~&~~
\infer[\oplus_r]{\dedllT{\Gamma,F\oplus G}}{\dedllT{\Gamma,G}} ~~&~~
\infer[\exists]{\dedllT{\Gamma,\exists x.F}}{\dedllT{\Gamma,\subst{F}{t}{x}}}
\end{array}
$$}
\caption{A one-sided, dyadic proof system for linear logic}
\label{system_for_LL}
\end{table*}

\subsection{\ehhf}
\label{ehhf}

The language \ehhf\ \cite{Delzanno97,DM99} 
is an executable specification  language for modeling
concurrent and resource sensitive systems, based on the general purpose
specification logical language \forum\ \cite{Mil96b}. \ehhf\ is a
multiset-based logic combining features of extensions of logic  programming
languages like \lprolog, e.g.  goals with implication and universal
quantification, with  the notion of \emph{formulae as resources} at the basis
of linear logic. \ehhf\ uses a subset of linear logic connectives and a 
restricted class of formulae as defined later.
An \ehhf-program $P$ is a collection of multi-conclusion clauses of the
form:
\[
A_1\para\ldots\para A_n\lollo Goal,
\]
where the $A_i$ are atomic formulae, the linear disjunction 
$A_1\para\ldots\para{A_n}$ corresponds to the head of the clause and $Goal$ is
its body. Furthermore, $A\lollo B$ is a linear
implication.  Execution of clauses of this kind concurrently 
 {\em consumes} the resources (formulae)
they need in order to be applied in a resolution step.

Given a multiset of atomic formulae (the state of the computation) $\Omega_0$,
a resolution step $\Omega_0\rightarrow \Omega_1$  can be performed by applying
an instance $A_1\para\ldots\para A_n\lollo G$ of  a clause in the program $P$,
whenever  the multiset $\Theta$ consisting of the atoms  $A_1,\ldots,A_n$ is
contained in $\Omega_0$. $\Omega_1$ is then obtained  by removing $\Theta$ from
$\Omega_0$ and  by adding $G$ to the resulting multiset.  In the \ehhf\
interpreter, instantiation is replaced by unification.  At this point, since
$G$ may be a complex formula, the search rules  (i.e., the logical rules of the
connectives occurring in $G$) must be exhaustively applied in order to proceed.
Such a derivation corresponds to a specific branch of the proof tree of a multiset
$\Omega$. $\Omega$ represents the current global  state, whereas  $P$ describes
a set of clauses that can be triggered at  any point during a computation.

\ehhf\ provides a way to ``guard'' the application of a given clause.
In the extended type of clauses
\[
G_1~\with\ldots\with ~G_m~\intui (A_1\para\ldots\para A_n\lollo Goal),
\]
the goal-formulae $G_i$ are \emph{conditions} that must be solved in order
for the clause to be triggered. 

New components can be added to the current state by using goal-formulae of the
form $G_1\para G_2$. In fact, the goal  $G_1\para G_2,\Delta$ simply reduces to
$G_1,G_2,\Delta$. Conditions over the current state can be tested by using
goal-formulae of the form $G_1\with G_2$. In fact, the goal  $G_1\with
G_2,\Delta$ reduces to $G_1,\Delta$ and $G_2,\Delta$.  Thus, one of the two
copies of the state can be consumed to verify  a contextual condition.  
Universal quantification in a goal-formula $\forall x. G$ can be  used  to
create a new identifier $t$ which must be local to the derivation tree of the
subgoal $G[t/x]$. Finally, the constant $\all$ succeeds in any context and the
constant $\anti$ is simply removed from the current goal.

\subsubsection{Semantics}
\label{ehhf_sem}

The \ehhf\ operational semantics is given by means of 
a set of rules  describing the way sequents can be rewritten.
See \cite{DM99} for all details and for the results of correctness and
completeness w.r.t. linear logic. 
According to the {\em proof as computation interpretation} of linear logic, 
sequents represent the state of a computation.

Sequents assume the following form (simplified for the sake of presentation): 
$$ \Gamma; \Delta \rightarrow \Omega, $$
where $\Gamma$ and $\Delta$
are multi-sets of $\D$-formulae (respectively the unbounded and bounded context), 
and $\Omega$ is a multi-set of $\G$-formulae which contains the concurrent 
resources present in the state.
The class of formulae is restricted to two main classes, $\D$-
and $\G$-formulae ($D$ stands for {\em definite} clauses 
and $G$ for {\em goals}):

\mbox{
\small$
\hspace*{1cm}
\begin{array}{lll}
\\
\D ~~&::=&~~ \D\,\with\,\D~|~\forall x.\D~|~\Head\lollo\G~|
~\D\into G~|~\Head\\[0.2cm]
\G ~~&::=&~~ \G\,\with\,\G~|~\G\para\G~|~\forall x.\G 
         ~|~\D\lolli\G~|~\D\intui\G~|~\A~|~\anti~|~\all\\[0.2cm]
\Head ~~&::=&~~ \Head\para\Head~|~\A_r\\
\end{array}$}
\\

$\A$ represents a generic atomic formula, whereas $\A_r$ is a {\em rigid}
atomic formula, i.e., whose top-level functor symbol is a constant.

The rules of \ehhf\
are divided into {\em right rules} (or {\em search rules})
and {\em left rules}. Right rules are used to simplify goals in
$\Omega$ until they become atomic formulae. They define the behavior of the various
connectives: $\all$ is used to manage termination, $\anti$ to encode a null
statement, $\with$ to split a computation into two branches which share the
same resources, $\forall$ to encode a notion of {\em hiding}, $\para~$
to represent concurrent execution, $\intui$ and $\lolli$ to augment the
resource context (respectively, the unbounded and the bounded context).

Left rules define backchaining over clauses built with the connectives
$\into$ and $\lollo$. As described above, the rule for $\lollo$ is similar 
to \textsf{Prolog} rewriting,
except that multiple-headed clauses are supported; besides, a clause can be
reusable or not depending on which context it appears in. The rule for $\into$
allows to depart an independent branch in an empty context (this is often
useful to verify side conditions or make auxiliary operations).

\subsubsection{Implementation}

A working interpreter for \ehhf\ has been developed by Bozzano in 
\textsf{Lambda Prolog}, a language originally developed by 
Miller and Nadathur, which offers support  
for higher-order abstract syntax, a new and increasingly popular way 
to view the structure of objects such as formulae and programs.

The code of the \ehhf\ interpreter can be downloaded from
\cite{ehhfInterpreter}, where an example
 implementing the specification described in 
\cite{BDMMZ99b} is also downloadable.

\subsubsection{Extensions}
\label{ehhfExtensions}

In the MAS context, \ehhf\ has been used to specify an architecture based on the 
BDI ({\em Belief, Desires, Intentions} \cite{RG95}) approach \cite{BDMMZ99a}
and to verify the correctness of a MAS where agents were specified by means of 
event--condition--action rules \cite{BDMMZ99b}.

\ehhf\ has also been used to model object-oriented and deductive databases \cite{BDM97}
and object calculi \cite{BDLM00}.

\subsubsection{Example}

The \ehhf\ program for the seller agent may be as follows: 
\begin{itemize}
\item Seller's initial facts:
\begin{itemize}
	\item[] {\em min-price(orange, 1)}.
	\item[] {\em max-price(orange, 2)}. 
	\item[] {\em storing(orange, 1000)}.
	\item[] {\em seller-mailbox([])}.\\
We assume that every agent has a mailbox which all the agents in the system
can update by calling a {\em send} predicate. The mailbox
of the seller agent is initially empty (we are using \prolog\ syntax
for lists).
\end{itemize}
\item Seller's life cycle:
 \begin{itemize}
	\item[] {\em $\forall$  Message, OtherMessages. \\
\hspace*{.5cm} seller-mailbox([Message$|$OtherMessages]) $\para$\\
\hspace*{.5cm} seller-cycle $\lollo$ \\
\hspace*{1cm} manage(Message) $\para$  \\
\hspace*{1cm} seller-mailbox(OtherMessages) $\para$  \\
\hspace*{1cm} seller-cycle.
}\\
To satisfy the {\em seller-cycle} goal, the seller agent must have at least
one message in its mailbox. In this case, it consumes the 
{\em seller-mailbox([Message$|$ OtherMessages])} and {\em seller-cycle}
goals and produces the 
new goals of  managing the received message ({\em manage(Message)}), 
removing it from the mailbox 
({\em seller-mailbox(OtherMessages)}, where the list of
messages does not contain {\em Message} any more) and cycling
 ({\em seller-cycle}).
\end{itemize}
\item Seller's rules for managing messages:
	\begin{itemize}
	\item[] {\em $\forall$  Buyer, Merchandise, Req\_Amnt, Price. \\
\hspace*{.5cm} Old\_Amount $>=$ Req\_Amnt  $\with$ \\
\hspace*{.5cm} difference(Old\_Amount, Req\_Amnt, Remaining\_Amnt) $\with$\\
\hspace*{.5cm} max-price(Merchandise, Max) $\with$ Price $>=$ Max $\implies$\\
\hspace*{1cm} manage(contractProposal(Buyer, Merchandise, Req\_Amnt, Price)) $\para$ \\
\hspace*{1cm} storing(Merchandise, Old\_Amount) $\lollo$ \\ 
\hspace*{1.5cm} storing(Merchandise, Remaining\_Amount) $\para$ \\
\hspace*{1.5cm} ship(Buyer, Merchandise, Req\_Amnt, Price) $\para$ \\
\hspace*{1.5cm} send(Buyer, accept(seller, Merchandise, Req\_Amnt, Price)).
}\\
The goals before the $\implies$ connective are not consumed by the execution
of the rule: they 
are used to evaluate values ({\em difference(Old\_Amount, Req\_Amnt, 
Remaining\_Am\-nt)}),
to compare values ({\em Old\_Amount $>=$ Req\_Amnt} and {\em Price $>=$ Max})
and to get the value of variables appearing in facts that are not changed by the
rule ({\em max-price(Merchandise, Max)}). In this case, they succeed if
the conditions for shipping merchandise are met. The goals 
{\em  storing(Merchandise, Old\_A\-mount)} and 
{\em manage(con\-tract\-Pro\-po\-sal(Buyer, Merchandise, Req\_Amnt, Price))} 
 are consumed; they are rewritten in 
{\em storing(Merchandise, Remaining\_Amount)} (the information about stored  merchandise is 
updated), {\em ship(Bu\-yer, Merchandise, Req\_Amnt, Price)} (the required amount of 
merchandise is ship\-ped) and 
{\em send(Buyer, accept(seller, Merchandise, Req\_Amnt, Price)} (the message for informing
{\em Buyer} that its proposal has been accepted is sent).  
The {\em ship} predicate will be defined by some rules that we do not describe here. 
\item[]
\vspace*{.3cm} {\em $\forall$  Buyer, Merchandise, Req\_Amnt, Price. \\
\hspace*{.5cm} min-price(Merchandise, Min) $\with$ Price $<=$ Min $\implies$\\
\hspace*{1cm} manage(contractProposal(Buyer, Merchandise, Req\_Amnt, Price)) $\lollo$ \\
\hspace*{1.5cm} send(Buyer, refuse(seller, Merchandise, Req\_Amnt, Price).
}\\
If the proposed price is too low ({\em min-price(Merchandise, Min) $\with$ Price $<=$ Min})
the {\em Buyer}'s proposal is refused.
\item[]
\vspace*{.3cm}
 {\em $\forall$  Buyer, Merchandise, Req\_Amnt, Price. \\
\hspace*{.5cm}  storing(Merchandise, Old\_Amount) $\with$ Old\_Amount $<$ Req\_Amnt $\implies$\\
\hspace*{1cm} manage(contractProposal(Buyer, Merchandise, Req\_Amnt, Price)) $\lollo$ \\
\hspace*{1.5cm} send(Buyer, refuse(seller, Merchandise, Req\_Amnt, Price)).
}\\
If there is not enough merchandise stored,
the {\em Buyer}'s proposal is refused.
\item[]
\vspace*{.3cm}
 {\em $\forall$  Buyer, Merchandise, Req\_Amnt, Price. \\
\hspace*{.5cm} Old\_Amount $>=$ Req\_Amnt  $\with$ \\
\hspace*{.5cm} min-price(Merchandise, Min) $\with$ Price $>$ Min $\with$\\
\hspace*{.5cm} max-price(Merchandise, Max) $\with$ Price $<$ Max $\with$\\
\hspace*{.5cm} eval-means(Max, Price, Means) $\implies$\\
\hspace*{1cm} manage(contractProposal(Buyer, Merchandise, Req\_Amnt, Price)) $\lollo$ \\
\hspace*{1.5cm} send(Buyer, \\
\hspace*{1.5cm} contractProposal(seller, Merchandise, Req\_Amnt, Means)).
}\\
If there is enough merchandise and the price proposed by {\em Buyer} is between 
the minimum and maximum prices established by the seller, the means of
{\em Price} and {\em Max} is evaluated ({\em eval-means(Max, Price, MeanPrice)})
and a {\em contractProposal} with this new price is sent to {\em Buyer}.
	\end{itemize}
\end{itemize}

\section{A Comparison among the Specification Languages}
\label{spec-logic-comp}
In this section we compare the agent specification languages introduced so far along 
twelve dimensions whose choice is mainly driven by \cite{creatingReusing}.  
Although it is difficult to assess if the following twelve dimensions are 
all and the only ones relevant for  characterizing an agent programming language,
we think that they represent a reasonable choice.  

In Section \ref{dimensions} 
we introduce the twelve dimensions. For each one
we explain why it is relevant for characterizing
an agent programming language.
We also formulate some questions whose answers, given in Section \ref{answers}, help
in understanding how each of the six languages
analyzed in this paper supports the given dimension.

\subsection{Comparison Dimensions}
\label{dimensions}

This paper is mainly concerned with the prototyping stage rather than
with the development of a final application. For this reason we avoid discussing all those
technical details which are not necessary for modeling, verifying and prototyping 
a MAS, such as efficiency, support for mobility and physical distribution, 
support for integration of external packages.
Indeed, we concentrate on dimensions related with 
the basic definition of an agent quoted in the introduction 
\cite{jennings:98} (dimensions 2, 3, 4, 5),
on dimensions related with the agent representation and management of
knowledge (dimension 6),
on dimensions related with the ability of a set of agents to form a MAS (dimensions 7, 8, 9),
and on dimensions which, although not peculiar of an agent programming language, 
are particularly important for the correct development of agents and a MAS (dimensions
10, 11, 12).

\begin{enumerate}
\item {\em Purpose of use.} Understanding
in which engineering/development stage the language proves useful
is necessary to adopt the right language at the right time.
\begin{itemize}
\item Is the language suitable for running autonomous agents 
in a real environment? 
\item Is the language suitable for MAS prototyping? 
\item Is the language suitable for verifying properties of the 
implemented MAS?
\end{itemize}
\item {\em Time.} Agents must both react in a timely fashion to actions taking place
in their environment and plan actions in a far future, thus they should be aware of
time.
\begin{itemize} 
\item Is time dealt with explicitly in the language? 
\item Are there operators for defining complex timed expressions? 
\end{itemize}
\item {\em Sensing.} One of the characterizing features of an agent is its ability 
to sense and perceive the surrounding environment.
\begin{itemize}
\item Does the language provide constructs for sensing actions 
(namely, actions which sense the environment)?
\end{itemize}

\item {\em Concurrency.} 
Agents in a MAS execute autonomously and concurrently and
thus it is important that an agent
language provides constructs for concurrency among agents (external concurrency) and 
concurrency within threads internal to the agent (internal concurrency).
\begin{itemize} 
\item Does the language allow the modeling of concurrent actions within 
the same agent? 
\item Does it support concurrency among executing agents?
\end{itemize}
\item {\em Nondeterminism.} 
The evolution of a MAS consists of a
nondeterministic succession of events. 
\begin{itemize}
\item Does the language support nondeterminism? 
\end{itemize}

\item {\em Agent knowledge.} The predominant agent model attributes human-like 
attitudes to agents. The agent knowledge is often characterized by beliefs, 
desires and intentions 
\cite{RG95}. Often, human beings are required to 
reason in presence of incomplete and uncertain knowledge.
\begin{itemize}
\item Does the language support a BDI-style architecture?
\item Does the language support incomplete agent knowledge?
\item Does it support uncertainty?
\end{itemize}

\item {\em Communication.} Agents must be 
social, namely, they must be able to 
communicate either with other agents and with human beings. 
\begin{itemize}
\item Are communication primitives provided by the language? 
\item Is it necessary for an agent to know details of another agent's implementation
in order to communicate with it, or does communication take place on a more
abstract level?
\item Is the programming language tied to some specific agent communication language?
\end{itemize}  
\item {\em Team working.} The ability to form team is becoming more and more
important in the intelligent agents research area as witnessed by the increasing 
number of researchers which address this specific topic\footnote{For example, during
the Second International Joint 
Conference on Autonomous Agents and Multiagent Systems (AAMAS) which took place in Melbourne 
in July 2003 an entire session was devoted to team working, with four papers presented; 
a large number of documents on team work is published by 
the TEAMCORE Research Group at the University of Southern California \cite{teamcore}.}. 
Building a team may involve
coordination/negotiation protocols.
\begin{itemize}
\item Is the language suitable for defining and programming teams? 
\item Is the language suitable for expressing coordination/negotiation protocols? 
\end{itemize}
\item {\em Heterogeneity and knowledge sharing.}
In many real systems agents are heterogeneous since they were developed by
different organizations with different (sometimes opposite) 
purposes in mind.
\begin{itemize} 
\item Which are the necessary conditions that agents must respect to interact?
\item Do agents need to share the same ontology? 
\item Are agents able to cope with the heterogeneity of information?
\end{itemize}

\item {\em Programming style.}
The language programming style may be more or less suitable for implementing  
a given reasoning mechanism or a given agent architecture.
\begin{itemize}
\item Does the language support goal-directed reasoning, forward reasoning, reactiveness?
\item Does the agent programming language require to stick to a fixed agent model or does 
it leave the choice to the programmer? 
\end{itemize}
\item {\em Modularity}.
Agent programs are typically very complex and a developer
would benefit from structuring them by defining modules, macros and procedures. 
\begin{itemize}
\item Does the language provide constructs for defining modules, macros and/or
procedures? 
\end{itemize}
\item {\em Semantics.} Due to the complexity of languages for agent, providing 
a clear semantics is the only means to fully understanding the meaning of 
the constructs they 
provide and thus exploiting the potentialities of the language. 
\begin{itemize}
\item Is a formal semantics of the language defined? 
\item Are there 
results explaining the link between system execution and formal semantics? 
\end{itemize}
\end{enumerate}

\subsection{The Six Languages Compared along the Twelve Dimensions}
\label{answers}

\paragraph*{Purpose of use.}
\congo\ allows the design of flexible controllers for agents living in complex scenarios.
Its extension \textsf{IndiGolog} provides a practical framework for 
real robots that must sense the environment and react to changes
occurring in it, and \textsf{Legolog} is an agent 
architecture for running \textsf{IndiGolog} high-level programs 
in \textsf{Lego MINDSTORM} robots.
\textsf{CASL} \cite{shapiro-lesperance-levesque}) is an environment based on
\congo\ which provides a verification environment.

\agento\ is suitable for modeling agents and MAS. We are not aware of papers
on the suitability of \agento\ or its extensions 
for verifying MAS specifications or implementing real 
agent systems.  

\impact's main purpose is to allow the integration of heterogeneous information sources
and software packages. It has been used to develop real applications 
ranging from  combat information management where \impact\ 
was used to provide yellow pages matchmaking services
to aerospace applications where \impact\ technology has led to the 
development of a multiagent solution to the ``controlled flight into terrain'' problem.
The \textsf{IADE} environment provides support for monitoring the MAS evolution.

\dylog\ is suitable for building 
agents acting, interacting and planning in dynamic environments.
A web agent system called \textsf{WLog} has been developed using
 \dylog\ to demonstrate \dylog's 
potential in developing adaptative web applications as software agents.

In \cite{fisherICTL94} a range of sample applications of \concMTM\
utilizing both the core features of the language and some of its extensions are discussed. 
They include bidding, problem solving, process control, fault tolerance.
\concMTM\ has the potential of specifying 
and verifying applications in all of the areas above \cite{ijcis97}, but
it is not suitable for the development of real systems.

\ehhf\ can be used for MAS modeling and verification, as discussed in
\cite{BDMMZ99b}. It is not suitable for running autonomous agents 
in a real environment.

\paragraph*{Time.}
In \congo\ time instants correspond directly with situations: 
$s_0$ is the agent's situation at time $0$, $do([a_1, ..., a_n], s_0)$ is the 
agent's situation 
at time $n$. We can think of a succession of situations as a discrete time line.
Most temporal modalities as found in temporal logics can be expressed in situation
calculus using quantification over situations.

In \agento\ time is included in all the constructs
of the language. The operations allowed on time variables are only
mathematical operations (sums and differences). When programming
an agent, it is possible to specify the time grain of its execution. 

Time is a central issue in \concMTM\ specifications: there are a lot of
time-based operators (``{\em since, until, in the next state, in the last state, sometime in 
the past, sometime in the future, always in the past, always in the future}'') which 
allow the definition of complex timed expressions.

As far as the other languages are concerned,  time does not appear in 
expressions of the language, either explicitly or implicitly. 

\paragraph*{Sensing.}
\dylog\ is the only language which provides an explicit construct for defining 
actions which sense the value of a fluent. However  all the languages allow perception of
values of atoms that are present in their knowledge base. Whether this knowledge base
correctly maintains a model of the environment or not, and thus whether it is possible 
to ``sense'' the surrounding environment or not, depends on the given specification. 
In our running example, all the agents maintain the information about the 
stored amount of oranges locally in their knowledge bases. In practice 
this information should be obtained by physically sensing the
environment (the warehouse, in this case), since nothing ensures that 
the agent's information is consistent with the environment state. 

It is worthwhile to note that, in a certain sense, the \impact\ agent programming
language is the only one which really senses its (software) environment
by means of the code calls mechanism: this mechanism allows an agent to
get information by accessing external software packages.   

We also note that, although the \congo\ language does not support
sensing primitives, \textsf{IndiGolog} does, and that 
sensing in the situation calculus is discussed by \cite{Reiter01}.

\paragraph*{Concurrency.} 
\congo\ provides different constructs for concurrent execution of
processes; these processes may be either internal to a single agent or may 
represent different agents executing concurrently. Thus, \congo\ supports both
concurrency of actions inside an agent and concurrency of agents.

The same holds for \ehhf, where it is possible to concurrently 
execute either goals internal to a single agent or goals for activating
different agents. As an example of the last case, if different agents
were characterized by a {\em cycle} like the one depicted for the seller agent, 
it would be possible to prove a goal like
{\em agent1-cycle} $\|$ {\em agent2-cycle} $\|$ ... $\|$ {\em agentN-cycle} 
meaning that {\em agent1} to {\em agentN} are executed concurrently.

As far as \impact\ is concerned, it associates a body of code implementing a notion 
of concurrency to each
agent in the system, to specify how concurrent actions internal to the agent must be
executed. Concurrency among agents cannot be explicitly specified. 

The converse situation
takes place with \concMTM, where concurrency of internal actions is not supported;
a \concMTM\ specification defines a set of concurrently executing agents which
are not able to execute internal concurrent actions.

Both \dylog\ and \agento\ do not support concurrency at the language level. 

\paragraph*{Nondeterminism.} 
\congo\  allows for nondeterministic choice between actions, 
nondeterministic choice of arguments and nondeterministic iteration.

Nondeterminism in the \impact\ language derives from the fact that
the feasible, rational and reasonable status sets giving the semantics 
to agent programs are not unique, thus introducing nondeterminism in the
agent's behavior. 

In \dylog\ and \ehhf\ nondeterminism is introduced, as in 
usual logic programming settings, by the presence of more procedures 
(rules, in \ehhf) defining the same predicate. 

The main source of nondeterminism in \concMTM\ is due to nondeterministic temporal
operators such as ``sometime in the past'', ``sometime in the future'', 
which do not identify a specific point in time, but may be
verified in a range of time points. 

\agento\ does not seem to support any kind
of nondeterministic behavior.

\paragraph*{Agent knowledge.}

A \congo\ model can include the specification of the agents' mental states,
 i.e., what knowledge and goals 
they have, specified in a purely declarative way \cite{levesqueLNAI1441}. 
With respect to incomplete knowledge, 
\begin{quote}
\congo\ can accommodate 
incompletely specified models, both in the sense that the initial 
state of the system is not completely specified, and in the sense that the 
processes involved are nondeterministic and may evolve in any number of ways
\cite{lesperanceAOIS99}. 
\end{quote}

\agento\ allows for expressing beliefs, capabilities, commitments, obligations. 
It does not allow the representation of intentions, goals and desires and 
it does not support reasoning mechanisms in presence of incomplete knowledge or
uncertainty. \textsf{PLACA} adds intentions and plans to the data structures provided
by \agento.

Beliefs of \impact\ agents consist of the values returned by the 
packages accessed by the agent by means of the code call mechanism. No
intentions and desires are ascribed to \impact\ agents: an 
\impact\ agent is characterized by its obligations, permissions, prohibitions.
Two extensions of basic \impact\ agent programs, namely 
probabilistic and meta agent programs,  can deal with uncertainty and
beliefs about other agents beliefs, respectively. 

Beliefs of \dylog\ agents are represented by the values of the 
functional fluents characterizing the agent's knowledge. These values range
over  true, false and unknown. The ``unknown'' value
allows \dylog\ agents to reason in presence of incomplete knowledge, as discussed 
by \cite{reasoningWithIncompleteKnowledge}. Agents can perform
hypothetical reasoning on possible sequences of actions by exploring different
alternatives.
Recent extensions allow \dylog\ agents to represent beliefs of
other agents in order to reason about conversation protocols
\cite{reasoningAboutConvProt,reasoningAboutSelfAndOthers}.

The beliefs of \concMTM\ agents in a given time point
consist of the set of predicates true in that time point.  
Adding deliberation and explicit beliefs to \concMTM\ agents is discussed in
\cite{michaelIJCAI97} and the extension of \concMTM\ agents with confidence
is dealt with in \cite{michaelAAMAS02}. 

In \ehhf\ beliefs are represented by true facts.
Goals are neither explicitly represented nor maintained 
in persistent data structures during the agent execution: 
they are managed in the usual way in a logic programming 
setting.
\ehhf\ does not provide language constructs for representing desires, 
intentions and other mental attitudes, but it may be 
adopted to model a BDI architecture \cite{BDMMZ99a}.

\paragraph*{Communication.} 
The specification of communicative multiagent systems in \congo\ is discussed in
\cite{levesqueLNAI1441}. A meeting scheduler multiagent system example is
used to show in practice the proposed approach.

Among \agento\ language constructs, there are 
the {\em INFORM, REQUEST} and {\em UNREQUEST} communicative actions which constitute a
set of performatives upon which any kind of communication can be built. 
Communication in \agento\ is quite rigid since, for  agent $A$ to request 
an action to  agent $B$ it is necessary 
to know the exact syntax of the requested action. The receiver agent has no
means to understand the content of a request and  perform an 
action consequently, if the  action to be performed is not exactly specified as 
the content of the message itself.
This is clearly a strong limitation, which recent agent communication
languages, such as \kqml\ \cite{Mayfield:95} and \textsf{FIPA ACL} \cite{fipa-acl} 
have partially addressed. The integration of \agento\ and \kqml\ 
proposed in \cite{agentk} aims at making communication management in
\agento\ more flexible. 

The same limitation affecting \agento\ also affects
\concMTM: every agent has a communicative
interface which the other agents in the system must know in order to exchange information.
Despite this limitation, \concMTM\ has been proposed both as a coordination
language \cite{michaelCOORDINATION97} and as a language for forming groups of agents 
where agents have the ability to broadcast messages to the members
of a group \cite{michaelISLIP00,michaelATAL98}.

The \impact\ language does not provide communication primitives as part of the language,
but among the software packages an agent may access there is a {\em msgbox} package
providing message box functionalities. Messages can have any form, adhering to some
existing standard or being defined ad-hoc for the application.

In \cite{reasoningAboutSelfAndOthers} \dylog\ agents are 
enriched with a communication kit
including a primitive set of speech acts, a set of special ``get message'' actions and
a set of conversation protocols. Exchanged messages are based on \cite{fipa-acl}.

An agent behavior driven by the reception of a message and
the verification of a condition on the current state can be easily modeled in \ehhf.
In \cite{BDMMZ99b} the translation of rules
``{\em on receiving MessageIn check Condition update State send MessageOut}''
 into \ehhf\ is shown. Messages can have any form.

\paragraph*{Team working.}
The attention devoted to teamwork in a MAS setting is quite recent. 
For this reason, none of the languages discussed so far encapsulates 
explicit constructs for team specification and programming.
The developer can define protocols for forming teams and she/he can try to
program agents which respect the given protocol. According to the 
support given to communication (previous paragraph), the task of defining such
protocols may be more or less difficult. 

With respect to the six languages we analyzed in this paper, 
the only papers explicitly addressing the problem of forming groups 
are \cite{michaelISLIP00,michaelATAL98} dealing with flexible grouping in \concMTM. 
In \concMTM, 
a group is essentially a set consisting of both agents and further sub-groups. 
The basic properties of groups are that agents are able to broadcast 
a message to the members of a group, add an agent to a group, 
ascertain whether a certain agent is a member of a group, 
remove a specified agent from a group, and construct a new subgroup.

We are not aware of similar extensions of 
\congo, \agento, \impact, \dylog\ and \ehhf.

\paragraph*{Heterogeneity and knowledge sharing.}

Agents programmed in the same 
language {\em have the potential} to interact without respecting 
any specific condition. Whether the agents {\em will} interact or not
depends on the correctness of their code with respect to 
the specification of the application. Whatever the language used is, if agent {\em A}
sends  a message {\em Message} to agent {\em B} and the code of agent
{\em B} does not include rules (or imperative statements, or
clauses) for managing {\em Message}, {\em A} and {\em B}
will not be able to engage in a dialog. 

Among the six languages discussed in this paper, \impact\ is the most 
suitable one to cope with 
heterogeneity of data sources. Any information source or software application
can be accessed through an \impact\ program, provided it is
properly ``agentified''. \impact\ can be seen as a programming
layer providing a uniform access to heterogeneous sources of information and
applications. 

The other five languages are not conceived for accessing heterogeneous data 
sources and for integrating the information contained in the data sources: they provide no
support for these two tasks.

In all of the six languages, agents {\em may} take advantage of sharing
the same ontology to interact,
but they {\em are not forced} to do so. 
To make an example, \dylog\ agents can refer to a common ontology contained in the
domain knowledge. This approach is described in
\cite{ApplyingLogicInferenceTechniques}.
However, it is also possible to develop \dylog\ agents without 
explicitly defining a common ontology. 
When developing a MAS, the developer has in mind 
the ontology the agents will refer to. Although it is a good practice to 
make it explicit, this is not compulsory to guarantee the MAS working.

\paragraph*{Programming style.}

The constructs provided by \congo\ allow 
to mimic both goal-directed reasoning  (``{\em if the current goal is G 
then call the procedure to achieve G}'')
and reactiveness (``{\em interrupt as soon as the condition C becomes true}'').

\agento\ programming style is reactive: 
depending on the message received by the agent and on its current beliefs, 
a commitment rule can be used.

\impact\ implements a forward reasoning mechanism: the interpreter looks for all
the action status atoms which are true with respect to the current state, 
the agent program and the agent integrity constraints.   

For \dylog\ a goal directed proof procedure is defined, which allows to compute a query 
from a given dynamic domain description. 

\concMTM\ is defined as a language for modeling reactive systems \cite{concMTM13}.
Thus, reactiveness is the predominant feature of \concMTM\ agents.

\ehhf\ supports a Prolog-like goal-directed reasoning mechanism.

All of the six languages are flexible enough to specify/implement
agents adhering to different agent models.

\paragraph*{Modularity.} 
All the languages described in this paper support modularity at the agent level, since
they allow the definition of each agent program separately from the definition of the 
other agents.

\congo\ and \dylog\ both support the definition of procedures. 
In \congo\ these procedures are defined by macro expansion into formulae of the 
situation calculus, while in \dylog\ they are defined as axioms in the dynamic
modal logic.

\agento\ does not support the definition of procedures, even if in Section 
6.3 of \cite{shoham93}  macros are used for readability sake. The macro expansion
mechanism is not supported by the \agento\ implementation. 

\ehhf\ supports the definition of procedures as logic programming languages do, by
defining rules for solving a  goal. 

Finally, \impact\ and \concMTM\
do not allow the definition of procedures.

\paragraph*{Semantics.}
All the languages discussed in this survey, except for \agento, have a formal 
semantics. 

Semantics of \congo\ is given as a transition semantics
by means of the predicates $Final(\delta, s)$ and 
$Trans(\delta, s, \delta', s')$. The possible configurations that can be reached 
by a program $\delta$ in situation $s$ are
those which are obtained by repeatedly following the transition relation 
starting from $(\delta, s)$ and which are final.
Different interpreters for languages extending or 
slightly modifying \congo\ have been 
proven correct with respect to the intended semantics of the language. 
See for example \cite{DLL},
\cite{adaptingGolog} and \cite{extendingAnswerSetPlanning}.

There are three different semantics which can be associated with an
\impact\ agent program, given its current state and integrity constraints: 
the feasible, rational and reasonable status set semantics.
Reasonable status set semantics is more refined than the rational one, which is more
refined than the feasible one. All of them are defined as a set of action status
atoms of the form  $\bfDo\alpha(\vect)$
that are true with respect to the agent program $\agprog$, the current state $\agstate$
and the set $\IC$ of underlying integrity constraints. 
In \cite{eite-subr-99} algorithms for evaluating
the semantics of arbitrary agent programs are proposed and their complexity
is evaluated; computing the 
reasonable status set semantics of a proper subset of \impact\ 
agent programs, called regular agents, is possible in polynomial time, as
demonstrated in \cite{eite-etal-00a}.
 
The  logical characterization of \dylog\
is provided in two steps. First, a multimodal logic 
interpretation of a dynamic domain description which describes the monotonic part 
of the language is introduced. Then, an abductive semantics to 
account for non-monotonic behavior of the language is provided.
\dylog\ is defined by a  proof procedure which is sound and complete with respect to
the Kripke semantics of modal logic.

The semantics of \concMTM\ is the one defined for 
the first-order temporal logic FML. It is a Kripke-style semantics given by 
the $\models$ relation that assigns the truth value
of a formula in a model ${\cal M}$ at a particular moment in time $i$ and
with respect to a variable assignment. The soundness, completeness and 
termination of the resolution procedure discussed in \cite{resolutionMethodTL}
have been established.

The \ehhf\ operational semantics is given by means of 
a set of rules  describing the way sequents can be rewritten.
According to the {\em proof as computation interpretation} of linear logic, 
sequents represent the state of a computation. Soundness and completeness 
results are established with respect to linear logic.

\section{Related Work}
\label{related-and-future}

To the best of our knowledge, there are only few previous attempts to analyze 
and compare a large set of logic-based formalisms and calculi for 
multiagent systems.

Some issues such as Kripke models and possible world semantics, Shoham's
\agento, \textsf{Concurrent METATEM} 
etc. are briefly surveyed in \cite{WJ95} but not specifically 
in a logic-based perspective.

A discussion on the adoption of logic programming and 
non-monotonic reasoning for evolving knowledge bases (and, more in general,
for intelligent agents) can be found in \cite{leite}. The book focuses 
on logic programming for non-monotonic reasoning and on languages for updates.
It analyzes and compares \textsf{LUPS} \cite{lups}, 
\textsf{EPI} \cite{epi} and introduces their extensions
\textsf{KUL} and \textsf{KABUL}. For some of the languages discussed
in \cite{leite} a working interpreter exists: see 
\cite{updatesURL} for details.
We will shortly describe \textsf{LUPS} and \textsf{KABUL} in the sequel as
representative examples of languages of updates. 

The work which shares more similarities with ours is the paper 
``{\em Computational Logic and Multi-Agent Systems: a Roadmap}''  \cite{CL-and-MAS}.
That paper discusses  different formalisms with respect to 
the representation of the agent's mental state, the agent life-cycle, 
the ability to communicate following complex interaction protocols and 
the capability of representing and reasoning about other agents' beliefs. 
The languages and systems analyzed in Sadri and Toni's
roadmap include \textsf{INTERRAP} \cite{ST57,ST45,interrap},
\textsf{3APL} \cite{ST41,goaldirected3apl}, and the work by
\cite{kowa-sadr-99}, \cite{ST67},  \cite{ST2}, 
 \cite{ST60}, \cite{ST76}, \cite{ST24}, \cite{ST25}, \cite{ST23},
\cite{ST42}, \cite{ST10} and \cite{ST62}.
Our paper complements Sadri and Toni's survey because, apart from 
the \impact\ language and part of the work on \caselp\ which are also 
discussed by Sadri and Toni, we analyze different 
languages and approaches from different perspectives. 
Sadri and Toni mainly aim at putting in evidence the contribution of
logic to knowledge representation formalisms and to basic mechanisms
and languages for agents and MAS modeling. Our paper analyzes a subset of
logic-based executable languages whose main features are their suitability for
specifying agents and MASs and their possible integration in the 
\arpeggio\ framework. 
We think that a researcher interested in 
logic-based approaches to multiagent systems modeling and prototyping can find 
an almost complete overview in reading both Sadri and Toni's paper 
and ours.

Other relevant logic-based approaches that are dealt with 
neither in \cite{CL-and-MAS} nor in this work are
\textsf{ALIAS}, \textsf{LUPS},
\textsf{KABUL}, \textsf{AgentSpeak(L)} and the \textsf{KARO} framework.
Given more time and space, they could fit in 
the picture and in the future we would like to check the feasibility of 
incorporating some of them in the \arpeggio\ framework.\\

\textsf{ALIAS} ({\em Abductive LogIc AgentS} \cite{alias4}) is an agent architecture 
based on intelligent and social logic agents where the main form of agent reasoning 
is abduction. \textsf{ALIAS} agents can coordinate their reasoning with 
other agents following several coordination schemas. In particular, they can either 
cooperate or compete in the solution of problems. 
The \textsf{ALIAS} architecture is characterized by two separate layers:
the lower layer involves reasoning while the upper layer involves 
social behavior.

Agent reasoning is specified by abductive logic  programs consisting of
a set of clauses {\em Head :- Body}, where
{\em Head} is an atom and 
{\em Body} is a conjunction of literals (atoms and negation of atoms), 
plus a set of abducible 
predicates and a set of integrity constraints. 

Agent interaction and coordination
is specified in  a logic language named 
\textsf{LAILA} ({\em Language for AbductIve Logic Agents}) suitable for modeling
agent structures from the viewpoint of social behavior.  
\textsf{LAILA} provides high-level declarative operators such as 
the communication operator $>$, the competition operator $;$, the collaboration
operator $\with$ and a down-reflection operator $\downarrow$ which allows a local 
abductive resolution to be triggered. 
The operational semantics of \textsf{LAILA} is discussed in 
\cite{alias4}. 

The distinguishing features of \textsf{ALIAS} consist of its support for 
(i) \emph{coordinating the reasoning} of agents 
at a high level of abstraction, (ii) explicitly referring to the
\emph{hypothetical reasoning} capabilities of agents from within the language, 
and (iii) providing the tools to maintain the system (or part of it) \emph{consistent}
with respect to the integrity constraints.

A prototypical version of \textsf{ALIAS} has been implemented on top of
\textsf{Jinni} \cite{jinni}, a logic programming language extended with primitives
for concurrent programming.

The main difference between \textsf{ALIAS} and 
the six languages discussed in our paper lies
in the background logic upon which the languages are based: \textsf{ALIAS}
is based on first-order logic while all the six languages discussed 
in this paper encapsulate features of linear, modal or temporal extensions 
of first-order logic. 

\textsf{ALIAS} shares with \textsf{Concurrent METATEM} and
\agento\ the support for communication at the language level.
However, the communication
primitives provided by \textsf{ALIAS}, associated with a semantics of collaboration
and competition and with the local abductive reasoning operator
(down-reflection), are more expressive than those provided by 
\concMTM\ and \agento\
and more suitable for tackling problems that require the coordination of
agent reasoning in presence of incomplete knowledge.
 
The agent reasoning form (abduction) is a shared feature between 
\textsf{ALIAS} and \dylog. \\

The language \textsf{LUPS} ({\em ``the language for  dynamic updates''} \cite{lups}) 
is based on a notion of update commands
that allow the specification of logic programs. Each command in \textsf{LUPS} can
be issued in parallel with other \textsf{LUPS} commands and specifies an update action, 
basically encoding the assertion or retraction of a logic program rule. 
An extension to \textsf{LUPS}, \textsf{EPI} ({\em ``the language around''} \cite{epi}), 
introduces the
ability to access external observations and make the execution of programs dependent on
both external observations and concurrent execution of other commands. 
\textsf{EVOLP} ({\em EVOlving Logic Programs} \cite{evolp}) 
integrates in a simple way the concepts of both Dynamic Logic Programming
 and \textsf{LUPS}.\\

\textsf{KABUL} ({\em Knowledge And Behavior Update Language} 
\cite{leite}) overcomes some limitations of \textsf{LUPS}. 
In particular, it allows the specification of 
 updates that depend on a 
sequence of conditions (``assert a rule {\em R} if some condition {\em Cond1} 
is true after some condition {\em Cond2} was true''), 
delayed effects of actions,
updates that should be executed only once,
updates that depend on the concurrent execution of other commands,
updates that depend on the presence or absence of a specific rule in the
knowledge base, and
inhibition of a command.
Moreover, with respect to \textsf{LUPS}, \textsf{KABUL} provides 
more flexible means to specify updates that will occur in the future and to deal with
effects of actions.\\

\textsf{AgentSpeak(L)} \cite{agentspeakl} is a programming language based on a restricted
first order language with events and actions. The behavior of the agent is dictated
by the programs written in \textsf{AgentSpeak(L)}. The beliefs, desires and intentions of
the agent are not explicitly represented as modal formulae. The current state of the
agent, which is a model of itself, its environment and other agents, is viewed as its current
belief state; states which the agent wants to bring about based on its external or internal 
stimuli can be viewed as desires; and the adoption of programs to satisfy such stimuli can 
be viewed as intentions. An operational semantics of 
\textsf{AgentSpeak(L)} is provided, as well as the proof theory of the language.
\textsf{AgentSpeak(XL)} \cite{agentspeakxl} integrates a task scheduler into
 \textsf{AgentSpeak(L)} to ensure an efficient intention selection.
The paper \cite{agentsp3apl} demonstrates that every agent which can be programmed
in \textsf{AgentSpeak(L)} can be programmed in the already cited 
\textsf{3APL} language. The authors of \cite{agentsp3apl} write that, in their opinion, 
the converse (simulating \textsf{3APL} by  \textsf{AgentSpeak(L)}) is not feasible 
and thus they conjecture that \textsf{3APL} has strictly more expressive power than 
\textsf{AgentSpeak(L)}. A simulation of \congo\ by \textsf{3APL} 
has also been provided \cite{congo3apl}
showing that \textsf{3APL} and \congo\ are closely related languages. \\

The framework \textsf{KARO} ({\em Knowledge, Abilities, Results and Opportunities}
\cite{karo}) formalizes motivational attitudes situated at two different levels. At the 
{\em assertion} level (the level where operators deal with assertions), {\em preferences}
and {\em goals} are dealt with. At the {\em practition} level (the level where operators
range over actions) {\em commitments} are defined. The main informational attitude 
of the \textsf{KARO} framework is {\em knowledge}. The fact that agent  $i$ knows
$\varphi$ is represented by the formula {\bf K}$_i \varphi$ and is interpreted in a 
Kripke-style possible worlds semantics.
At the action level {\em results}, {\em abilities} and {\em opportunities} 
are considered. The abilities of an agent are formalized via the {\bf A}$_i$ operator:
{\bf A}$_i \alpha$  denotes the fact that agent $i$ has the ability to do $\alpha$. 
Dynamic logic is used to formalize the notions of opportunities and results.
$do_i(\alpha)$ refers to the performance of action $\alpha$ by the agent $i$.
The formula $\langle do_i(\alpha)\rangle\varphi$ represents the fact that
agent $i$ has the opportunity to do $\alpha$ and that doing $\alpha$ leads to $\varphi$.
The formula
$[do_i(\alpha)]\varphi$
states that if the opportunity to do $\alpha$ is indeed present, doing $\alpha$ 
results in $\varphi$. 
Starting from these basic attitudes, preferences, goals and commitments 
can be modeled. Different methods for realizing automated reasoning within 
agent-based systems modeled using the \textsf{KARO} framework are discussed
in \cite{HustadtDixonEtal01a,HustadtDixonEtal01b}.

\section{Conclusion}
\label{spec-logic-conc}

In this paper we have systematically analyzed and compared six logic-based
and executable MAS specification languages. 
Although these languages were chosen on the basis of their
potential to be integrated in the \arpeggio\ framework,
they are an interesting and representative set of formalisms based
on extensions of first order logic. 
We have discussed the logic-based formalisms upon which the languages are
built to allow the reader to understand the theoretical 
foundations of the languages. We have demonstrated
the use of these languages by means of an example, and we have  
compared them along twelve dimensions. Finally, we have surveyed
other approaches adopting computational logic for MAS specification.

Various advantages in using logic-based approaches for modeling and prototyping
agents and MAS should emerge from this paper:
as pointed out by Wooldridge and Jennings in  Section 2 of \cite{WJ95}
agents are often modeled in terms of mental attitudes such as beliefs, 
desires, intentions, choices and commitments. 
The main advantage in using logic and modal logic in particular 
for modeling intentional systems is 
that it allows to easily and intuitively represent intentional notions without 
requiring any special training.
The {\em possible world} semantics usually adopted for modal languages has 
different advantages: it is well studied and well understood, and the associated 
mathematics of ``correspondence theory'' is extremely elegant.
Formal languages that support temporal operators are a powerful means for 
specifying sophisticated {\em reactive} agents in a succinct fashion. 
Moreover, since agents are expected to {\em act}, languages based on 
formal theories of action such as dynamic logic and the situation calculus 
are extremely suitable to model agents' ability to perform actions. 

According to the observations above, if we compare logic languages and 
object-oriented formalisms for the specification of agents 
we note that logic languages are more suitable than object-oriented 
languages to model agents.  
According to \cite{odell2002} autonomy and interaction are the key features
which differentiate agents and objects.
Autonomy has two independent aspects: dynamic autonomy and 
nondeterministic autonomy. Agents are dynamic because they can exercise 
some degree of activity, rather than passively providing services. 
With respect to dynamic autonomy, agents are similar to active objects.
By means of the running example we have shown that logic-based languages
are suitable for expressing the active behavior
of agents in a concise and simple way. 
Agents may also employ some degree of unpredictable (or nondeterministic) 
behavior. We have shown that all of the six languages analyzed in this paper
support some kind of nondeterminism. 
The ``or'' connective and the ``exists'' quantifier
introduce a degree of nondeterminism to all the 
languages based on first order logic. 
Interaction implies the ability to communicate with the environment and other 
entities. Object messages (method invocation) can be seen as the most basic
form of interaction. 
A more complex degree of interaction would include those agents that 
can react to observable events within the environment. 
And finally in multiagent systems, agents can be engaged in multiple, 
parallel interactions with other agents. Logic-based languages
prove their suitability in modeling agents that react to an 
event (logical implications of the form {\em if the event E took place 
then something becomes true} can be used for this purpose) and to 
reason about sophisticated conversations.

The last consideration of our paper deals with the implementation of a MAS prototype:
we have seen that different languages among the ones we
discussed have an interpreter which extends logic programming in some way.  
Using a logic programming language for MAS prototyping has different advantages:
\begin{itemize}
\item {\em MAS execution}: the evolution of a MAS consists of
a nondeterministic succession of events; from an abstract point of view
a logic programming language is a nondeterministic language in which computation 
occurs via a search process.
\item {\em Meta-reasoning capabilities}:  
agents may need to dynamically modify their behavior so as to
adapt it to changes in the environment. Thus, the possibility given by
logic programming of viewing programs as data is very important in this setting.
\item {\em Rationality and reactiveness of agents}:
the {\em declarative} and the {\em operational}
interpretation of logic programs are strictly related to the main
characteristics of agents, i.e., {\em rationality} and {\em
reactiveness}.
In fact, we can think of a {\em pure} logic program as the specification of
the rational component of an agent and
we can use the operational view of logic programs (e.g.
left-to-right execution, use of non-logical predicates)
to model the reactive behavior of an agent.
The adoption of logic programming for combining reactivity and rationality is
 described in \cite{Kowalski:96}.
\end{itemize}

\section*{Acknowledgements}

We want to thank Marco Bozzano (ITC - IRST, Trento, Italy) and 
Giorgio Delzanno (University of Genova, Italy) 
for their  useful suggestions. We also thank 
Matteo Baldoni (University of Torino, Italy), 
for his clarification on some aspects related with \dylog\ and Paolo Torroni 
(University of Bologna, Italy)
for his support in the description of \textsf{ALIAS} and \textsf{LAILA}.
Finally we thank the anonymous referees for their comments which helped to improve the 
paper.

\label{lastpage}

\end{document}